\documentclass[lettersize,journal]{IEEEtran}
\usepackage{amsmath,amsfonts}
\usepackage{algorithmic}
\usepackage{algorithm}
\usepackage{array}
\usepackage[caption=false,font=normalsize,labelfont=sf,textfont=sf]{subfig}
\usepackage{textcomp}
\usepackage{stfloats}
\usepackage{url}
\usepackage{verbatim}
\usepackage{graphicx}
\usepackage{cite}
\usepackage{tabularx}
\usepackage{multirow}
\usepackage{bm}
\usepackage{marvosym}
 
\usepackage{amsmath, amsfonts, amsthm, stackrel, amssymb} 
\usepackage{colortbl}  %彩色表格需要加载的宏包
\usepackage{xcolor}
\usepackage{array}   %对表列和表格线的设置需要
\begin{document}

\title{FreeQ-Graph: Free-form Querying with Semantic Consistent Scene Graph for 3D Scene Understanding}

\author{Chenlu Zhan, Yufei Zhang, Gaoang Wang ~\IEEEmembership{Member,~IEEE}, Hongwei Wang~\IEEEmembership{Member,~IEEE}
        % <-this % stops a space
\thanks{This work was supported in part by Zhejiang Provincial Natural Science Foundation of China (LDT23F02023F02) and  the National Natural Science Foundation of China (No.62106219). \emph{Corresponding Authors: Hongwei Wang and Gaoang Wang.}}% <-this % stops a space
\thanks{Chenlu Zhan is with the College of Computer Science and Technology, Zhejiang University, Zhejiang, China (chenlu.22@intl.zju.edu.cn)}
\thanks{Yufei Zhang is with the College of Biomedical Engineering and Instrument Science, Zhejiang University, Zhejiang, China (yufei1.23@intl.zju.edu.cn)}
\thanks{Gaoang Wang, and Hongwei Wang are with Zhejiang University-University of Illinois Urbana-Champaign Institute, Zhejiang University, Haining, China. (gaoangwang@intl.zju.edu.cn, hongweiwang@intl.zju.edu.cn) 
}
}

\markboth{Journal of \LaTeX\ Class Files,~Vol.~14, No.~8, August~2021}%
{Shell \MakeLowercase{\textit{et al.}}: A Sample Article Using IEEEtran.cls for IEEE Journals}


\maketitle
\begin{abstract}

%Understanding semantic interactions in complex 3D scenes through free-form language presents a significant challenge.
Semantic querying in complex 3D scenes through free-form language presents a significant challenge.
Existing 3D scene understanding methods use large-scale training data and CLIP to align text queries with 3D semantic features. However, their reliance on predefined vocabulary priors from training data hinders free-form semantic querying.
Besides, recent advanced methods rely on LLMs for scene understanding but lack comprehensive 3D scene-level information and often overlook the potential inconsistencies in LLM-generated outputs.
In our paper, we propose \textbf{FreeQ-Graph}, which enables \textbf{Free}-form \textbf{Q}uerying with a semantic consistent scene \textbf{Graph} for 3D scene understanding. 
The core idea is to encode free-form queries from a complete and accurate 3D scene graph without  predefined vocabularies, and to align them with 3D consistent semantic labels, which accomplished through three key steps.
We initiate by constructing a complete and accurate 3D scene graph that maps free-form objects and their relations through LLM and LVLM guidance, entirely free from training data or predefined priors.
Most importantly, we align graph nodes with accurate semantic labels by leveraging 3D semantic aligned features from merged superpoints, enhancing 3D semantic consistency.
To enable free-form semantic querying, we then design an LLM-based reasoning algorithm that combines scene-level and object-level information to intricate reasoning.
We conducted extensive experiments on 3D semantic grounding, segmentation, and complex querying tasks, while also validating the accuracy of graph generation. Experiments on 6 datasets show that our model excels in both complex free-form semantic queries and intricate relational reasoning.
\end{abstract}
%(c) Our “free-form” approach, leveraging LLM and LVLM agents, extends from limited vocabulary to free-form queries while utilizing a 3D semantic-consistent scene graph to precisely identify objects and their relationships.}}
%Comparison of complex semantic querying between our FreeQ-Graph, designed for free-form querying, and previous SOTA 3D open-vocabulary semantic segmentation methods.}}
%\label{fig:0}
\begin{figure}[h]
\centering
\includegraphics[width=1\linewidth]{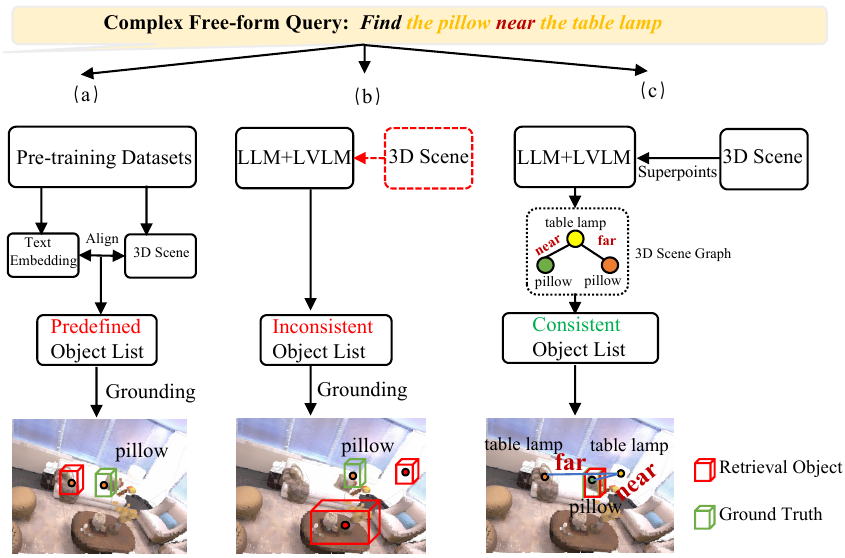}
\caption{We introduce FreeQ-Graph, a 3D scene understanding work for free-form complex semantic querying with a semantic consistent scene graph. (a) The open-vocabulary methods depend on pre-trained data and predefined objects to align text with 3D features, limiting their support for free-form queries.
%The CLIP-based  ``open vocabulary" approaches rely on pre-trained datasets to align textual queries with 3D scene features, inherently depending on a limited set of predefined category labels. Such reliance restricts their ability to support free-form semantic querying and relational reasoning.
(b) %Some methods over rely on LLMs and LVLMs for reasoning, but their lack of 3D scene awareness often results in object lists misaligned with true 3D semantics, causing inaccurate conclusions.
Some methods overly depend on LLMs and LVLMs for reasoning, yet their lack of 3D scene awareness often yields object lists misaligned with actual 3D semantics, leading to inaccurate reasoning.
%Recent LLM-based querying methods rely entirely on the reasoning capabilities of LLMs and LVLMs, yet overlook their lack of 3D scene-level information. As a result, the generated object lists often deviate from the actual 3D semantics, leading to inaccurate reasoning outcomes.
(c) We propose a training-free framework that leverages LVLMs and LLMs to build a complete 3D spatial scene graph for free-form querying without predefined priors. Superpoint merging ensures the alignment of 3D node features with correct semantic labels, enabling accurate and consistent 3D scene understanding.}
%Our work adopt a training-free framework using LVLM and LLM to construct a complete 3D spatial scene graph to map free-form queires without any priors. By merging superpoints, it aligns accurate 3D semantic features, enabling precise and consistent free-form semantic querying.}
%Our free-form approach adopts a training-free paradigm based on LVLM and LLM, constructing a complete 3D spatial scene graph to substantially reduce hallucinated or inaccurate nodes generated by the models. Furthermore, by integrating direct 3D scene information through superpoint merging, we ensure consistent alignment with accurate 3D semantic features. This design enables precise semantic querying in a free-form manner.}
\label{fig:0}
\end{figure}


\section{Introduction}
\label{sec:intro}
Interacting with 3D scenes using open-vocabulary perception is a key challenge facing current AI-driven agents~\cite{xiong20253ur,wu2025language}. Accurately querying semantic objects and their relationships through complex free-form queries in intricate 3D environments remains an unresolved issue~\cite{peng2023openscene7}.

Recent works~\cite{ding2023pla9,shafiullah2022clip,yang2023llmgrounderopenvocabulary3dvisual} tackling 3D scene understanding tasks frequently rely on CLIP~\cite{zhang2023clip} to align textual queries with scene semantics, which heavily depends on large-scale pre-trained datasets.
%Recent works~\cite{ding2023pla9,shafiullah2022clip,yang2023llmgrounderopenvocabulary3dvisual} addressing 3D scene understanding tasks often rely on CLIP~\cite{zhang2023clip} to fuse visual and textual modalities. They leverage CLIP to encode open-vocabulary queries, aligning them with relevant 2D text fragments, thereby serving as 3D base models for new 3D data.
%However, these models are limited by CLIP’s bag-of-words limitations, lacking capabilities for complex text queries, visual reasoning, and spatial awareness~\cite{yang2023llmgrounderopenvocabulary3dvisual}.
Meanwhile, some methods leverage large language models (LLMs)~\cite{achiam2023gpt,yang2023llmgrounderopenvocabulary3dvisual,fu2024scenellm} to facilitate flexible semantic interactions, which are crucial for handling complex queries in 3D scene understanding.
%Meanwhile, large language models (LLMs)~\cite{achiam2023gpt} exhibit strong language understanding and support flexible, free-form text interactions, which are essential for complex queries in 3D scene understanding.
We aim to develop a training-free framework that can  acquire semantically aligned 3D features to support accurate free-form querying in 3D scenes. However, existing methods still encounter significant limitations: \textbf{1) Limited predefined vocabulary priors from training datasets hinder free-form semantic querying.} 
%We aim to build 3D semantic-aligned scenes to facilitate free-form querying and reasoning, yet current methods~\cite{schult2023mask3d,jatavallabhula2023conceptfusion12,koch2024open3dsg} face significant challenges: \textbf{1) Limited predefined vocabulary priors from training datasets hinder free-form semantic querying.} 
Most 3D scene understanding models~\cite{cai20223djcg,jia2024sceneverse,koch2023sgrec3dselfsupervised3dscene} depend on large-scale training data and use CLIP to encode queries and scenes, inherently constraining them to a fixed set of predefined categories. This limitation hinders their capacity for free-form semantic querying and relational reasoning, as illustrated in Fig.~\ref{fig:0} (a).
%Most 3D scene understanding models \cite{kerr2023lerf,kobayashi2022decomposing} leverage large-scale data for training and rely on CLIP for encoding queries and scenes. This inherently ties querying to a limited set of predefined category labels. Such reliance restricts their ability to support free-form semantic querying and relational reasoning, as shown in Fig.~\ref{fig:0} (a).
%confines them to the finite vocabulary in training priors, limiting their ability to handle free-form queries and reasoning, as shown in Fig.~\ref{fig:0} (a).
%The CLIP-based  ``open vocabulary" approaches rely on pre-trained datasets to align textual queries with 3D scene features, inherently depending on a limited set of predefined category labels. Such reliance restricts their ability to support free-form semantic querying and relational reasoning.
%Most 3D scene understanding models~\cite{kerr2023lerf,kobayashi2022decomposing} rely on CLIP to encode text and image scenes, which restricts them to ``bag-of-words" representations, rendering them incapable of handling complex text queries and relational reasoning.
\textbf{2) Inconsistency between 3D instance features and semantic labels.}
Recent methods~\cite{gu2023conceptgraphsopenvocabulary3dscene41,linok2024barequeriesopenvocabularyobject42,yang2023llmgrounderopenvocabulary3dvisual,fu2024scenellm} rely solely on LLMs and LVLMs to generate semantic labels for 3D instance features, yet neglect the lack of 3D scene information. This often leads to inconsistent or incorrect outputs, where objects and relations misalign with true 3D semantics, resulting in unreliable reasoning, as shown in Fig.~\ref{fig:0}(b).
%\textbf{2) Inconsistency between 3D scene instance features and their semantic labels.} Existing methods~\cite{gu2023conceptgraphsopenvocabulary3dscene41,linok2024barequeriesopenvocabularyobject42,yang2023llmgrounderopenvocabulary3dvisual,fu2024scenellm} rely entirely on LLMs and LVLMs to generate semantic labels for 3D scene instance features, yet overlook their lack of 3D scene-level information and ignore potentially inconsistent outputs generated by LLMs. As a result, the generated object lists and relations often deviate from the actual 3D semantics, leading to inaccurate reasoning outcomes, as illustrated in Fig.~\ref{fig:0} (b). %struggling with aligning 3D point representations with their correct semantic labels, as illustrated in Fig.~\ref{fig:0} (b).
%Recent
%LLM-based querying methods rely entirely on the reasoning capabilities of LLMs and LVLMs, yet overlook their lack of 3D scene-level information. As a result, the generated object lists often deviate from the actual 3D semantics,leading to inaccurate reasoning outcomes. 
%State-of-the-art methods utilize extensive textual descriptions and meticulously labeled datasets for training and fine-tuning, but they achieve good results only for objects and queries annotated in the training set. This limits their ability to freely generalize to complex language and relational queries, constraining their understanding of 3D open scenes.
\textbf{3) Lack of scene spatial relation reasoning.} 
Current methods~\cite{yin2024sai3d,nguyen2024open3dis} predominantly focus on object-level segmentation and retrieval, while disregarding spatial relationships within complex scenes. This oversight substantially constrains their capability to handle semantic relationship queries.
%Current methods~\cite{yin2024sai3d,nguyen2024open3dis} focus on object-level segmentation and retrieval, overlooking spatial relationships within complex scenes. This limits the ability to query semantic relationships.
%Current mainstream approaches predominantly rely on LLMs to predict a limited set of objects and their descriptive labels, neglecting scene-level information. This oversight hampers the retrieval and reasoning of smaller objects and the overall context of 3D scenes.

In our paper, we propose FreeQ-Graph, a training-free framework that enables free-form semantic querying with a semantic consistent scene graph for 3D scene understanding.
Our key innovation lies in a training-free, free-form querying framework that constructs a scene graph with accurate nodes and relations, aligns 3D instances with correct semantics through superpoint merging, and integrates LLM-based reasoning for spatial queries, setting our approach apart.
%Our key innovation is a training-free free-form querying framework that builds a scene graph with accurate nodes and relations, aligns 3D instances with correct semantics via superpoint merging, and incorporates LLM-based reasoning for spatial queries, distinguishing our approach. %Our key innovation is a training-free free-form querying framework that constructs a scene graph with accurate nodes and relations, with semantic alignment that unifies 3D instance nodes with correct semantic labels, along with LLM-based reasoning for scene spatial querying, setting our approach apart.
1) We construct a complete and accurate 3D scene graph using LVLMs and LLMs to map free-form instances and their relationships, without relying on any training priors.
Unlike ConceptGraph~\cite{gu2023conceptgraphsopenvocabulary3dscene41}, which depends on 2D models and often misses or duplicates objects, our approach ensures accurate scene representation through mutual correction between agents and the grounded model.
2) We align free-form nodes with consistent semantic labels to obtain 3D semantically consistent representations. This is achieved by generating superpoints and performing structural clustering to extract 3D instance features and their semantic labels, thereby aligning each 3D point with its corresponding semantics. In contrast, others~\cite{gu2023conceptgraphsopenvocabulary3dscene41} struggle with a consistent semantic representation.
%for unifying point representations and nodes, while others~\cite{gu2023conceptgraphsopenvocabulary3dscene41} struggle with it. 
%This module enriches each node's semantic features with semantic aligned features derived from merged super points, aligning free-form nodes with their semantic representations.
3) We develop an LLM-based reasoning algorithm that breaks complex queries into CoT-reasoning by combining scene and object-level information for free-form querying. In contrast, other~\cite{gu2023conceptgraphsopenvocabulary3dscene41} single strategy lacks scene context, limiting its query capabilities.
We conduct thorough experiments on six datasets, covering 3D semantic grounding, segmentation, and complex querying tasks, while also validating the accuracy of scene graph generation. The results demonstrate that our model excels in handling complex semantic queries and relational reasoning.
%e conducted challenging experiments in 3D semantic segmentation, retrieval, and complex semantic queries, demonstrating that our LLM-based, vocabulary-free 3D scene understanding model outperforms CLIP mapping in complex scene comprehension. This underscores our model's significant superiority in understanding 3D scenes regarding complex textual queries, visual relational reasoning, and spatial awareness.
Our contributions are summarized as follows:
\begin{itemize}
\item We propose FreeQ-Graph, a training-free free-form querying framework with the semantic consistent scene graph and LLM-based reasoning algorithm for 3D scene understanding.

\item We propose a 3D semantic alignment method that aligns 3D graph nodes with consistent semantic labels, enabling the extraction of free-form 3D semantic-aligned features.
%semantic alignment to unify the node and point representation
%scene graph for free-form semantic interaction, equipped with a semantic point-node alignment module to extend the flexibility of consistent semantics.
%the semantic structure and subordinate relationships of complex scenes.
\item We introduce an LLM-based CoT-reasoning algorithm that combines scene-level and object-level information for scene spatial reasoning.

\item Extensive experiments on 6 datasets demonstrate that our method excels in querying complex free-form semantics and relation reasoning perceptions.
\end{itemize}

\section{Related Work}
\label{sec:formatting}

\begin{figure*}[h]
\centering
\includegraphics[width=0.96\linewidth]{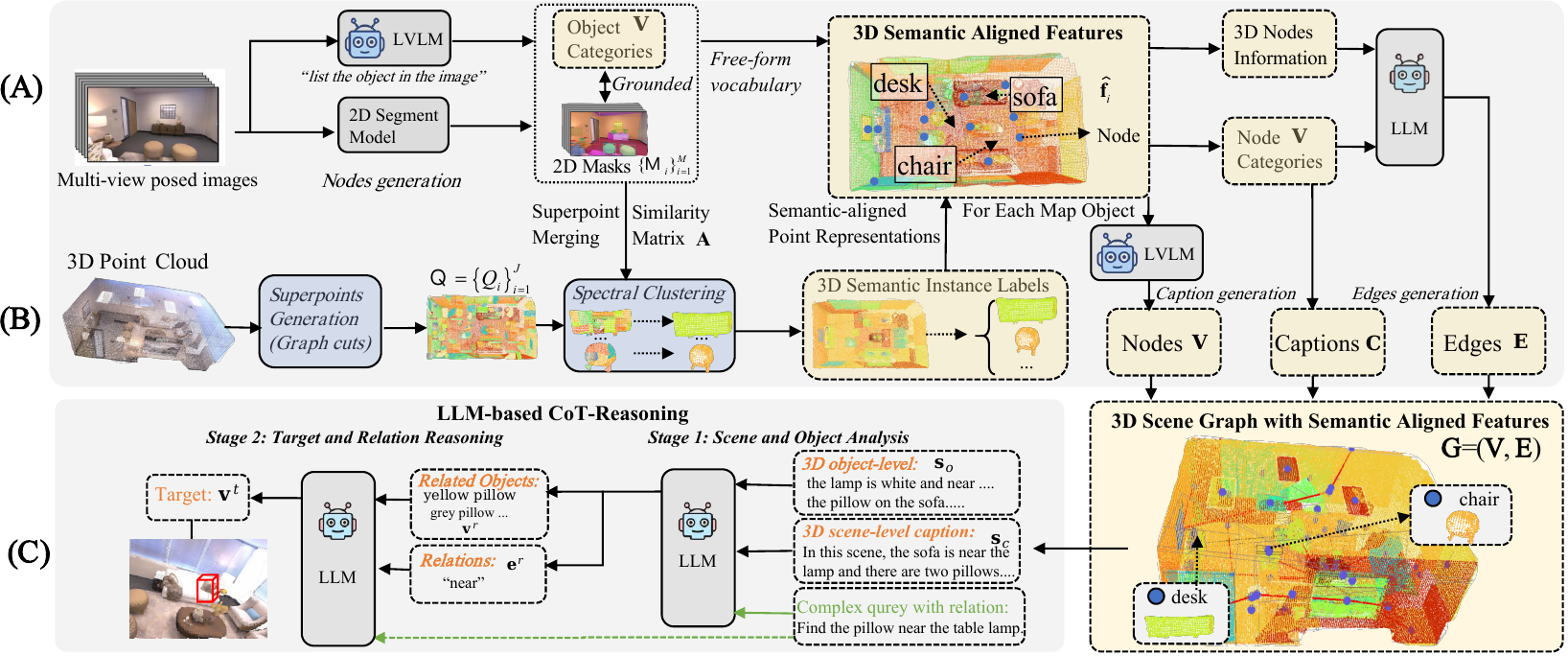}
\caption{{The structure of  FreeQ-Graph.} We propose FreeQ-Graph to realize free-form semantic querying without any training priors for 3D scene understanding. 
(A) A 3D spatial scene graph to enable free-form semantic interaction with complete and accurate nodes and relations, (B) A semantic alignment module to align graph nodes with consistent semantic labels, and (C) An LLM-based CoT-reasoning algorithm for free-form semantic spatial querying.}
\label{main}
\end{figure*}

\subsubsection{Open-Vocabulary 3D Scene Understanding}
Natural language querying in complex 3D scenes demands deep understanding of free-form semantics and relationships. Many prior works~\cite{jia2024sceneverse,cai20223djcg,yuan2021instancerefer5,wu2023eda,yu2025inst3d,chen2022languageViL3DRel,wang2025liba,graphtmm2,3ddetecttmm} rely on joint training with large-scale pretrained data to align 3D scenes and query embeddings, but their dependence on predefined vocabularies limits true free-form querying. Recent advanced methods~\cite{linok2024barequeriesopenvocabularyobject42,gu2023conceptgraphsopenvocabulary3dscene41,yang2023llmgrounderopenvocabulary3dvisual,xu2024vlm,li2025seeground} leverage large language models (LLMs) for flexible semantic reasoning, yet they overly depend on LLMs and LVLMs to generate semantic labels for 3D features without sufficient 3D scene awareness~\cite{tmm4}. This often results in inconsistent or inaccurate outputs where objects and relations misalign with actual 3D semantics. Besides, some LLM-based methods~\cite{peng2023openscene7,huang2024chat,zhu2024scanreasonreground,yu2025inst3d} fine-tune on task-specific datasets, improving performance but still restricting free-form queries and requiring substantial training resources.
Our work utilizes a training-free, free-form querying
framework that constructs a scene graph with accurate nodes and relations, aligns 3D instances with correct semantics for scene understanding.

%LLM-Grounder~\cite{yang2023llmgrounderopenvocabulary3dvisual} decomposes complex natural language queries into semantic components with LLM, using tools like OpenScene~\cite{peng2023openscene7} or LERF~\cite{kerr2023lerf} to identify objects in 3D scenes, extending to novel scenes and arbitrary text queries. However, it still falls short in capturing 3D relational reasoning, attributes, and sequential semantics, and cannot overcome inherent LLM biases.
%For ours, we leverage the 3D semantic consistent scene graph containing the graph nodes with relations based on LLM and LVLM for free-form semantic queries, without any training priors.
%We leverage LLM and scene graphs to expand limited vocabulary queries into free-form semantic queries. We design a 3D scene graph embedding semantic associations, relational reasoning, and sequence attributes to capture complex scene semantics.

\subsubsection{3D Scene Graphs}
The 3D Scene Graph (3DSG) represents scene semantics compactly, with nodes for objects and edges for their relationships~\cite{armeni20193dscenegraphstructure34,Kim_202035,3DSSG202036,graphtmm}. Recent methods~\cite{rosinol2021kimeraslamspatialperception37,hughes2022hydrarealtimespatialperception38,graphtmm3} use 3DSG for 3D scene representations. 
These works~\cite{wang2023vl,koch2024open3dsg,armeni20193dscenegraphstructure34}, such as VL-SAT~\cite{wang2023vl}, construct scene graphs to model scene representations, but are constrained by the closed vocabularies from their training data, limiting their ability to support free-form semantic queries. More recent approaches like ConceptGraph~\cite{gu2023conceptgraphsopenvocabulary3dscene41} and BBQ~\cite{linok2024barequeriesopenvocabularyobject42} leverage LLMs to generate nodes and edges in scene graphs. However, their heavy reliance on LLM-generated outputs without incorporating 3D scene context often leads to inconsistent scene representations misaligned with actual 3D semantics.
Our approach constructs a semantically consistent 3D scene graph by first obtaining complete and accurate free-form nodes, then aligning them with correct semantic labels.

\section{Method}
In this section, we propose FreeQ-Graph, a framework that enables free-form querying with (A) a 3D spatial scene graph with complete nodes and relations to support free-form query, (B) the semantic alignment module to align nodes with the consistent semantic label, and (C) a LLM-based CoT-reasoning for scene spatial querying, as shown in Fig.~\ref{main}.

\subsection{Problem Formulation} 
\label{define}
Given each 3D scene $\mathbf{P}$ with multi-view posed RGB observations $\mathbf{I}=\left\{I_i\right\}_{i=1, \ldots, M}$ as input, where $M$ is the total number of images. The objective of free-form querying via 3D scene graph is to depict a semantic 3D scene graph $\mathbf{G}=(\mathbf{V},\mathbf{E})$ as the 3D scene representation, where $\mathbf{V} = {\{ {\mathbf{v}_j}\} _{j = 1,...,J}}$ denote the set of 3D objects and edges $\mathbf{E} = {\{ {\mathbf{e}_k}\} _{k = 1,...,K}}$ represents the relation between them.  $\mathbf{G}$ constitutes a structured representation of the semantic content of the 3D scene. Based on this semantic representation of the 3D scene $\mathbf{G}$, during the reasoning phase, it interacts and queries with the query $q$ and finally outputs a final target $\mathbf{v}$.
%We aim to represent the 3D point cloud $\mathbf{P}$ as a 3D scene graph representation $\mathbf{G}=(\mathbf{V},\mathbf{E})$, where $\mathbf{V} = {\{ {\mathbf{v}_i}\} _{i = 1,...,J}}$ denote the set of 3D objects and edges $\mathbf{E} = {\{ {\mathbf{e}_k}\} _{k = 1,...,K}}$ represents the relation between them.  

\noindent \textbf{Nodes.}
For each object $\mathbf{v}_i \in \mathbf{V}$, we characterize it as $\mathbf{v}_i=\{ \mathbf{p}_i, \mathbf{f}_i, \mathbf{c}_i, \mathbf{b}_i, {n}_i \}$, where ${{\mathbf{p}}_i} = \{ {\mathbf{x}_j}\} _{j = 1}^{{N_i}}$ is the pointcloud that contains $N_i$ points $\mathbf{x}_j$, $\mathbf{f}_i$ is the semantic feature, $\mathbf{c}_i$ is the node caption, $\mathbf{b}_i$ is the 3D bounding box, ${n}_i$ is the id of the node.
We denote the set of all object categories as 
$\mathcal V$.

\noindent \textbf{Edges.}
For each pair of nodes $\mathbf{v}_i,\mathbf{v}_j$, we denote the edge $\mathbf{e}_{ij}=\{ 
\mathbf{r}_{ij}, \mathbf{d}_{ij} \}$, where $\mathbf{r}_{ij}  \in \mathbf{E}$ is the relation label that provides the underlying rationale, $\mathbf{d}_{ij}$ is the Euclidean distance between centers of bounding boxes for $\mathbf{v}_i$ and $\mathbf{v}_j$.

%Each object $v_j$ is defined by a 3D point cloud $p_{v_j}$ with a bounding box $B=\{c, s\}$ and a semantic feature vector $f_{v_j}$, where $c=\left(c_x, c_y, c_z\right)$ is the center and $s=\left(s_x, s_y, s_z\right)$ is the size. 
%Following ConceptGraph~\cite{gu2023conceptgraphsopenvocabulary3dscene41}, 
%The 3D scene graph map is constructed by integrating each new frame $I_i$ into the current object set $\mathbf{V}_{i-1}$ and updating existing objects or generating new ones which are contained in the set $\mathcal{V}$.
%The 3D scene graph is updated by incorporating each new frame \( I_i \) into the object set \( \mathbf{V}_{i-1} \), refining existing objects, or adding new ones in \( \mathcal{V} \).  
The construction of the object set $\mathbf{V}$ and edge set $\mathbf{E}$ is in Sec.~\ref{graph11}.
For each object $\mathbf{v}_i$, we define the object-level information $\mathbf{s}_{o_i}$ as $\mathbf{s}_{o_i}=\{\mathbf{c}_i, \mathbf{b}_i, {n}_i \}$. For better reasoning, we define scene-level information $\mathbf{s}_{c}$ which represents the scene captions.
The detailed reasoning algorithm is in Sec.~\ref{reasoning1}.
 %The corresponding set of multi-view 2D images is denoted as $\left\{I_i\right\}_{i=1, \ldots, M}$. 
\subsection{3D Scene Graph with Complete Nodes and Edges.}
\label{graph11}
%To enable mapping free-form objects and capturing spatial relations in 3D scenes, we construct the 3D scene graph $\mathbf{G}$ in three main steps: 
    To facilitate the mapping of free-form objects and the capture of relations in 3D scenes, the crux lies in acquiring complete nodes, encompassing all small and thin objects, along with their complete captions, and edges that include detailed relations reflecting complex semantic connections. To achieve this, we construct the 3D scene graph $\mathbf{G}$ through three primary steps:
{1) Complete and accurate nodes generation without any training priors.} 
It is a  {free-form method without relying on predefined vocabulary}.
{2) 3D semantic consistent feature generation.}
{3) Edges and captions generation.} 

\noindent \textbf{Complete 3D scene nodes generation without priors.}
%To ensure complete and accurate 3D proposals, we employ a dual approach using LVLM-based 2D-3D  revision method. 
To obtain objects with semantic labels without predefined vocabulary, we first adopt a large vision-language model (LVLM)~\cite{liu2024visualLLAVA} to obtain the object set, then use a 2D instance segmentation model~\cite{ren2024grounded} to correct potential hallucinations, forming the set $\mathcal V$ of object categories, which can be denoted as:
\begin{equation}
   {\mathcal{V}, \{ {\mathcal{M}_i}\} _{i = 1}^{{M}}} = \bigcup_{i=1}^M {\phi (LVLM({I_i}))} 
\end{equation}
where $\phi$ is the 2D segment model, ${\mathcal{M}_i}$ is the mask set of image $I_i$.
Specifically, for each 2D image view $I_i$, we prompt the LVLM model like \emph{``please list all the central objects in the scene, 
focus on {smaller or overlooked objects}, and visual attributes, omitting background details."}. We use {specific prompts} to focus on smaller or overlooked objects. We then parse the response and obtain the initial object list for each image.
To reduce  potential hallucinations by the visual agent, we subsequently employ a 2D instance segmentation model~\cite{ren2024grounded} to ground all initial objects, identifying the final grounded object lists 
  and obtaining the corresponding 2D object mask set 
 $\{ {\mathcal{M}_i}\} _{i = 1}^{{M}}$, representing the candidate objects on the posed image $I_i$.
 We then construct the objects set $\mathcal{V}$ by retaining categories from the combined set.
 %After summarizing the multiple captions for each node, we obtain the final caption for each node. %In the Fig.~\ref{nodecaptiopn}, we present a randomly selected node, including one of its views and the corresponding caption for that node.
%

\noindent \textbf{3D semantic consistent feature generation.}
%Building on ConceptGraph~\cite{gu2023conceptgraphsopenvocabulary3dscene41}, 
We extract visual descriptors (CLIP~\cite{zhang2023clip}) from 2D masks. Additionally, we generate 3D instance semantic labels via superpoint clustering and align nodes with point-level semantic representations (see Sec.~\ref{align} for details). The final nodes $\mathbf{v}_i$ consist of point cloud $\mathbf{p}_{i}$, unit-normalized feature ${{\hat {\mathbf{f}_i}}} $, and 3D box $\mathbf{b}_i$.

\noindent \textbf{3D nodes caption generation.} For each posed image, building on ConceptGraph~\cite{gu2023conceptgraphsopenvocabulary3dscene41}, we generate node captions via LVLM+LLM: (1) prompt LVLM with \emph{``describe the central object"} at top-$n$ clean viewpoints for initial descriptions; (2) distill coherent captions $\mathbf{c}_i$ via LLM refinement. 

\noindent \textbf{3D scene edges generation.}
Building upon 3D nodes and captions, we establish spatial edges through the 3D information analyzed by LLM.
For each pair of nodes $\mathbf{v}_i,\mathbf{v}_j$, 
we compute pairwise similarity matrices via 3D bounding box IoU, then prune edges using Minimum Spanning Tree  optimization.
 Next, we query LLMs with node captions/coordinates (e.g., \emph{``What is the relationship between 1 and 2?"}) to extract spatial relations.
We also calculate the Euclidean distance $\mathbf{d}_{ij}$ between box centers. Thus we can generate the edge  $\mathbf{e}_{ij}=\{ 
\mathbf{r}_{ij}, \mathbf{d}_{ij} \}$. %for nodes $\mathbf{v}_i$ and $\mathbf{v}_j$.

\subsection{3D Scene Graph with Semantic Aligned Features}
\label{align}
LLM-generated 3D scene graphs often suffer from potential inconsistencies with actual 3D semantics due to the lack of 3D scene information. This deficiency can lead to reasoning errors, such as misalignment between the semantic labels of 3D instance features and the nodes in the scene graph. To address this, we generate 3D semantic labels for instance features, ensuring precise alignment between each point and its corresponding semantics, thereby rectifying node-semantic misalignment.
%To align nodes in the scene graph with correct semantic labels, we aim to generate 3D semantic labels of 3D instance features,  ensuring each point is aligned with its correct semantics. 
\begin{figure}[h]
\centering
\includegraphics[width=1\linewidth]{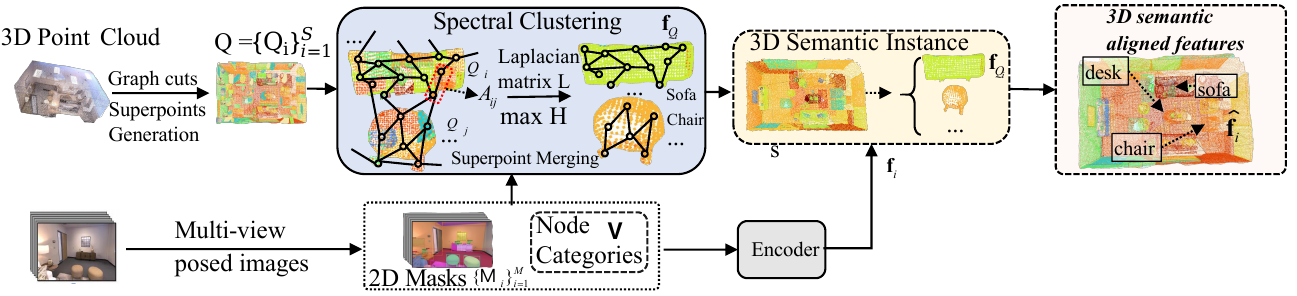}
\caption{Detailed process of 3D semantic instance label generation and semantic aligned features.}
\label{superpoint}
\end{figure}
This alignment occurs in two key steps: \textbf{(1) 3D semantic instance labels generation.} We apply graph cuts to segment the scene into superpoints, each superpoint represents a semantic label (e.g., ``desk" or ``chair"). Then we generate semantic labels through structural clustering for 3D instance features and align the graph nodes with consistent semantic labels. 
\textbf{(2) 3D semantic-aligned feature representation.} We integrate visual features with superpoint-based semantic label features to produce the final semantic-aligned features.

\noindent \textbf{3D semantic instance label generation.} 
We aim to generate 3D semantic labels of nodes from superpoint merging. 
Specifically, as shown in Fig.~\ref{superpoint}, inspired by PoLo~\cite{mei2024vocabulary}, we segment the 3D point cloud $\mathbf{P}$
  into $S$ superpoints $\mathcal{Q}=\left\{Q_i\right\}_{i=1}^S$ using graph cuts, where each $Q_i$ is a binary mask label of points.
  To merge superpoints into 3D instances, we construct a similarity matrix ${A}$, where each element $A_{ij}$ represents the similarity between superpoints ${Q}_i $ and ${Q}_j $: %incorporating both mask coherence and semantic coherence:
  \begin{equation}
\resizebox{0.90\hsize}{!}{$A_{ij} = \left( \sum_{m=1}^{M} g(O_{i,m}, \tau_{\text{iou}}) \cdot g(O_{j,m}, \tau_{\text{iou}}) \right) \cdot \frac{f_{Q_i}^\top f_{Q_j}}{\| f_{Q_i} \| \| f_{Q_j} \|}$}
 \end{equation}
 where $O_{i,m}$ and $O_{j,m}$ is the 2D mask projection of superpoint ${Q}_i$ and ${Q}_j$ in the $m$-th image. $g(O, \tau_{\text{iou}})$ is 1 if the IoU of mask $O$ exceeds threshold $\tau_{\text{iou}} $, and 0 otherwise.
$f_{Q_i}$ and $f_{Q_j}$ are the semantic representations of $Q_i$ and $Q_j$, obtained by encoding their label into feature vectors using a text encoder.
We then perform spectral clustering by the Laplacian matrix $L$ and segmenting superpoints via its eigenvectors. The optimal clustering dimension $H$ is set using the {eigengap heuristic}, selecting $H$ with the largest eigenvalue gap to determine the final number of superpoint semantic labels:
\begin{equation}
\resizebox{0.90\hsize}{!}{$ L = D^{-1/2} (D - A) D^{-1/2}, \quad
H = \arg\max_{1 \leq j \leq J-2} (\lambda_{j+1} - \lambda_j)$}
\end{equation}
where $D$ is the degree matrix, with $D_{ii} = \sum_j A_{ij}$. $\lambda_j$ are the eigenvalues of $L$, and the maximum gap corresponds to the optimal number of clusters.

\noindent \textbf{3D semantic aligned features.} 
After obtaining 3D semantic feature labels, we aim to align scene graph nodes with point-level semantic representations for 3D semantic aligned features. Unlike prior works~\cite{gu2023conceptgraphsopenvocabulary3dscene41,linok2024barequeriesopenvocabularyobject42} that rely solely on a visual encoder for per-point representation, we employ both a vision encoder and a text encoder to ensure semantic consistency between nodes and point representations. Specifically, 
%for each point $\mathbf{x}_j$ of the node $\mathbf{v}_i$ and its assigned semantic superpoint label $Q_i$, we first extract its visual feature 
for each the node $\mathbf{v}_i$ and its assigned semantic superpoint label $Q_i$, we first extract its visual feature 
$\mathbf{f}_i$ using CLIP. Additionally, we encode the superpoint’s semantic label $Q_i$ through a encoder to obtain its semantic feature ${{\mathbf{f}_{Q_i}}}$, and fuse it with the visual feature via meanpooling $\varphi$ to obtain the final semantically aligned representation ${\hat {\mathbf{f}_i}}$.

%We then use the obtained semantically consistent superpoint representations to enrich the 3D spatial graph representation. Specifically, for each point $\mathbf{x}_j$ of the node $\mathbf{v}_i$, we apply mean pooling $\psi$ with the corresponding points in the superpoint representation, thereby enhancing the semantic consistency of each node. Formally, it can be denoted as:
\begin{equation}
    %{{\hat {\mathbf{f}_i}}} = \varphi (\frac{1}{{{N_i}}}\sum\limits_{{\mathbf{x}_j} \in {{\mathbf{p}}_i}} {\psi ({{\mathbf{x}_j}})} ,{{\mathbf{f}_i}})
    {{\hat {\mathbf{f}_i}}} = \varphi ( {{\mathbf{f}_{Q_i}}},{{\mathbf{f}_i}})
\label{fi}
\end{equation}
%where $\psi$ is the superponts merging function and 
where $\varphi$ is the mean pooling. The ${{\hat {\mathbf{f}_i}}}$ is used to align semantic labels (e.g., “desk") to nodes in the graph. Errors in LLMs and LVLMs may misassign labels of 3D instance features, leading to incorrect results in JSON-based reasoning. 
%More details on superpoints can be found in {Appendix A.2}, and detailed samples are in {Appendix Listings 1-3}.
%More details on merging super points can be found in Appendix A.

\subsection{LLM-based CoT-Reasoning}
\label{reasoning1}
We designed the reasoning algorithm that breaks complex queries into CoT-reasoning, combining scene-level  $\mathbf{s}_c$ and object-level information  $\mathbf{s}_o$ (defined in Sec.~\ref{define}) for free-form semantic querying. Note that the CoT-reasoning are not separate. In the stage 1, we generate candidate objects, which are refined by further analysis in the next stage. 
%Detailed examples are in {Appendix B}.

\noindent \textbf{Stage 1: Scene and Object Analysis.}
As shown in Fig.~\ref{main} (C), to obtain the candidate targets $\mathbf{v}_r$ and relations $\mathbf{e}_r$, we input the user’s complex query $q$ alongside both object-level $\mathbf{s}_o$ and scene-level information $\mathbf{s}_{c}$ which defined in Sec.~\ref{define} into the LLM, which can be denoted as:
\begin{equation}
    n_r, \mathbf{e}_r = LLM(q,{\mathbf{s}_o},{\mathbf{s}_{\mathrm{c}}})
\end{equation}
where $n_r$ is IDs of candidate targets $\mathbf{v}_r$, scene-level information $\mathbf{s}_{c}$ represents scene captions.
This stage serves two purposes:
1) Leveraging the LLM’s planning ability to summarize observations of the entire scene, then decompose the complex semantic query into target and relational queries.
2) Cooperating object-level information with scene-level details, we aim to capture spatial relationships like ``near" in Fig.~\ref{main} (C), without overlooking smaller or less prominent objects.

The LLM agent decomposes the user’s query into object and relation queries, identifying candidate ids of related targets.
The ``object query" refers to the primary candidate objects cited in the semantic query.
%, with feedback on the volume of each candidate object. 
The ``relation query" identifies the relations of the candidate objects with the target, also providing the Euclidean distance of pairs.

\noindent \textbf{Stage 2: Target and Relation Reasoning.}
We further leverage the LLM for spatial reasoning based on the candidate objects $\mathbf{v}_r$ , relation $\mathbf{e}^r$, and the query $q$, then generate the final target $\mathbf{v}^t$. The reasoning stage can be denoted as:
\begin{equation}
    {\mathbf{v}^t} = LLM(q,\mathbf{v}_r,{\mathbf{e}^r})
\end{equation}
With the candidate IDs, we input the corresponding objects' captions, relations, 3D information, and the Euclidean distance from each candidate object to the centroid, along with the query $q$, into the LLM to infer the final target object.

\section{Experiment}
\subsection{Datasets and Implementation Details}
\subsubsection{Datasets}
We evaluated on Sr3D~\cite{achlioptas2020referit3d} and Nr3D~\cite{achlioptas2020referit3d}, and  ScanRefer~\cite{chen2020scanrefer} for visual grounding, and scene segmentation task on
Replica~\cite{replica19arxiv} and ScanNet~\cite{dai2017scannet} RGB-D. We validate the accuracy of scene graph on the 3DSSG dataset~\cite{3DSSG202036}.
\textbf{Sr3D}~\cite{achlioptas2020referit3d} dataset includes annotations based on spatial relationships between objects, while \textbf{Nr3D}~\cite{achlioptas2020referit3d} consists of human-labeled language object references. We selected a subset of 526 objects from Sr3D and filtered queries in Nr3D that only involved spatial relations between objects. For 8 corresponding ScanNet scenes, we conducted relational queries in the format (target, relation, anchor). We evaluate on Nr3D and Sr3D's standard splits using only the val set.
%, categorizing results by ``easy" and ``hard" cases, and metrics for ``view-dep." and ``view-indep."

\textbf{ScanRefer~\cite{chen2020scanrefer}} comprises 51,583 descriptions for 11,046 objects across 800 ScanNet~\cite{dai2017scannet} scenes. Following the benchmark, the dataset is split into train/val/test sets with 36,655, 9,508, and 5,410 samples, using the val set for evaluation. 
%“Fine-tuning” refers to methods that are adapted to specific tasks after pretraining, in contrast to those optimized through joint training across multiple tasks.
\begin{table*}[h]
	\centering
    \caption{Comparisons of 3D visual grounding on ScanRefer~\cite{chen2020scanrefer} dataset. The Accuracy at 0.25 and 0.5 IoU thresholds is presented separately for “Unique,” “Multiple,” and “Overall” categories. } 
	\scalebox{0.7}{
\small
	\setlength{\tabcolsep}{12pt}
\begin{tabular}{llcccccccc}
\hline
\multicolumn{1}{c}{\multirow{2}{*}{\textbf{Method}}} & \multicolumn{1}{c}{\multirow{2}{*}{\textbf{Venue}}} & \multirow{2}{*}{\textbf{Supervision}} & \multirow{2}{*}{\textbf{LLMs}} & \multicolumn{2}{c}{\textbf{Unique}}           & \multicolumn{2}{c}{\textbf{Multiple}} & \multicolumn{2}{c}{\textbf{Overall}} \\ \cline{5-10} 
\multicolumn{1}{c}{}                                 & \multicolumn{1}{c}{}                                &                                       &                                 & \textbf{Acc@0.25} & \textbf{Acc@0.5}          & \textbf{Acc@0.25}  & \textbf{Acc@0.5} & \textbf{Acc@0.25} & \textbf{Acc@0.5} \\ \hline
ScanRefer~\cite{chen2020scanrefer}                                              & ECCV’20                                             & Fully                                 & -                               & 67.60             & 46.20                     & 32.10              & 21.30            & 39.00             & 26.10            \\
InstanceRefer~\cite{yuan2021instancerefer5}                                          & ICCV’21                                             & Fully                                 & -                               & 77.50             & 66.80                     & 31.30              & 24.80            & 40.20             & 32.90            \\
3DVG-T~\cite{zhao20213dvg1}                                                 & ICCV’21                                             & Fully                                 & -                               & 77.20             & 58.50                     & 38.40              & 28.70            & 45.90             & 34.50            \\
BUTD-DETR~\cite{jain2022bottom}                                              & ECCV’22                                             & Fully                                 & -                               & 84.20             & 66.30                     & 66.30              & 35.10            & 52.20             & 39.80            \\
EDA~\cite{wu2023eda}                                                   & CVPR’23                                             & Fully                                 & -                               & 85.80             & 68.60                     & 49.10              & 37.60            & 54.60             & 42.30            \\
3D-VisTA~\cite{yang2023llmgrounderopenvocabulary3dvisual}                                               & ICCV’23                                             & Fully                                 & -                               & 81.60             & 75.10                     & 43.70              & 39.10            & 50.60             & 45.80            \\
G3-LQ~\cite{wang2024g}                                                 & CVPR’24                                             & Fully                                 & -                               & 88.60             & 73.30                     & 50.20              & 39.70            & 56.00             & 44.70            \\
MCLN~\cite{qian2024multi}                                                 & ECCV’24                                             & Fully                                 & -                               & 86.90             & 72.70                     & 52.00              & 40.80            & 57.20             & 45.70            \\
ConcreteNet~\cite{unal2024four}                                             & ECCV’24                                             & Fully                                 & -                               & 86.40             & 82.10                     & 42.40              & 38.40            & 50.60             & 46.50            \\
3DLFVG~\cite{zhang2024towards}                                                & CVPR’24                                             & Fully                                 & -                               & 65.80             & 51.27                     & 22.03              & 16.94            & 30.53             & 23.61            \\
3D-JCG~\cite{cai20223djcg}                                                 & CVPR’22                                             & Fully                                 & -                               & 83.47             & 64.34                     & 41.39              & 30.82            & 49.56             & 37.33            \\
Scene-Verse~\cite{jia2024sceneverse}                                            & ECCV’24                                             & Fully                                 & -                               & 81.60             & 75.10                     & 43.70              & 39.10            & 50.60             & 45.80            \\ 
TSP3D~\cite{guo2025text} & CVPR’25    &Fully      &  -   &  - & -  & -  & -  & 56.45  & 46.71 \\

LIBA~\cite{wang2025liba} & AAAI’25 &Fully       &     & 88.81  &74.27   & 54.42  & 44.41  &  59.57 &48.96 \\
\hline
WS-3DVG~\cite{wang2023distilling}                                                  & ICCV’23                                             & Weakly                                & -                               & -                 & -                         & -                  & -                & 27.40             & 22.00            \\ \hline
OpenScene~\cite{peng2023openscene7}                                              & CVPR’23                                             & Fine-tuning                           & CLIP                            & 20.10             & 13.10                     & 11.10              & 4.40             & 13.20             & 6.50             \\
Chat-3D v2 & NeurIPS’24                                            & Fine-tuning  &  Vicuna1.5-7B  &  61.20   &   57.60   &    25.20   &  22.60     & 35.90     & 30.40\\
ReGround3D~\cite{zhu2024scanreasonreground} & ECCV'24      & Fine-tuning      &FlanT5XL-3B     & - & -  & -  &  - & 55.50   &  50.20  \\   
ChatScene~\cite{huang2024chat} & NIPS'24      & Fine-tuning      &Vicuna1.5-7B     & - & -  & -  &  - & 55.50   &  50.20  \\   
Inst3D-LMM~\cite{yu2025inst3d} & CVPR’25    & Fine-tuning      &Vicuna1.5-7B     & 88.60  & 81.50  & 48.70  &  43.20 & 57.80  &  51.60 \\   
\hline
Open3DSG~\cite{koch2024open3dsg}                                               & CVPR’24                                             & Zero-Shot                             & CLIP                            & 17.54             & 11.38                     & 32.13              & -                & -                 & 14.32            \\
LERF~\cite{kerr2023lerf}                                                   & ICCV’23                                             & Zero-Shot                             & CLIP                            & -                 & -                         & -                  & -                & 4.80              & 0.90             \\
ConceptGraphs~\cite{gu2023conceptgraphsopenvocabulary3dscene41}                                                   & ICRA’24                                            & Zero-Shot                             & GPT-4                  & 16.50             & \multicolumn{1}{c}{10.32} & 9.57                 & 7.69                & 13.28                & 9.31                \\
BBQ~\cite{linok2024barequeriesopenvocabularyobject42}                                                   & Arxiv’24                                            & Zero-Shot                             & GPT-4o                   & 19.40             & \multicolumn{1}{c}{11.60} & -                  & -                & -                 & -                \\
LLM-Grounder~\cite{yang2023llmgrounderopenvocabulary3dvisual}                                                  & ICRA’24                                             & Zero-Shot                             & GPT-3.5                         & -                 & -                         & -                  & -                & 14.30             & 4.70             \\
LLM-Grounder~\cite{yang2023llmgrounderopenvocabulary3dvisual}                                                  & ICRA’24                                             & Zero-Shot                             & GPT-4 turbo                     & -                 & -                         & -                  & -                & 17.10             & 5.30             \\
ZSVG3D~\cite{yuan2024visual}                                                   & CVPR’24                                             & Zero-Shot                             & GPT-4 turbo                     & 63.80             & 58.40                     & 27.70              & 24.60            & 36.40             & 32.70            \\
VLM-Grounder~\cite{xu2024vlm} &CoRL’24 & Zero-Shot  & GPT-4o  & 66.00  &29.80  &48.30  &33.50       &  51.60  &32.80  \\
SeeGround~\cite{li2025seeground}                                                  & CVPR’25                                            & Zero-Shot                             & Qwen2-VL-72B                   & 75.70             & 68.90                     & 34.00              & 30.00            & 44.10             & 39.40            \\ 
%CSVG~\cite{li2025seeground}                                                  & CVPR’25                                            & Zero-Shot           &68.8& 61.2& 38.4 &27.3& 49.6 &39.8 \\
 
%Video-3d llm~\cite{zheng2025video} & CVPR’25    & Zero-Shot      &     &   &   &   &   &58.10   & 51.70  \\ 

\hline
%Chat-Scene~\cite{huang2024chat}                                             & NIPS'24                                             & Zero-Shot                             & Vicuna-7B-v1.5               & 89.59             & 82.49                     & 47.78              & 42.90            & 55.52             & 50.23            \\ \hline
%\textbf{Ours (GPT-3.5)        & 53.04 & 41.13 & 59.10 & 49.20 & \textbf{56.07} & \textbf{45.17} \\Ours (GPT-4)          & 53.46 & 41.86 & 59.30 & 49.70 & \textbf{56.38} & \textbf{45.78} \\Ours (GPT-4o)         & 54.13 & 42.41 & 59.80 & 50.10 & \textbf{56.97} & \textbf{46.26} \\Ours (Qwen2-VL-72B)   & 53.73 & 42.04 & 60.20 & 50.50 & \textbf{56.97} & \textbf{46.27} \\Ours (GPT-4 Turbo)    & \textbf{54.07} & \textbf{42.35} & \textbf{59.60} & \textbf{50.20} & \textbf{56.84} & \textbf{46.28} 
\textbf{Ours}                                             &    -                                                 & Zero-Shot                             & GPT-3.5                & 82.00 & 78.00 & 50.08 &  38.26 & 56.04 & 48.13     \\
\textbf{Ours}                                             &    -                                                 & Zero-Shot                    &GPT-4      & 82.10 & 78.70 & 50.82 &  37.98 & 55.46 & 48.86        \\
\textbf{Ours}                                             &    -                                                 & Zero-Shot           &  Vicuna1.5-7B           & 82.04 & 78.31 & 49.18 &  38.42 & 55.97 & 48.72          \\
\textbf{Ours}                                             &    -                                                 & Zero-Shot           &  Qwen2-VL-72B           & 83.10 & 79.40 & 50.16 &  39.13 & 56.13 & 49.41          \\
\textbf{Ours}                                             &    -                                                 & Zero-Shot                             &GPT-4 turbo               & 82.50 & 79.00 & 50.96 &  39.08 & 55.73 & 49.04          \\
\textbf{Ours}                                             &    -                                                 & Zero-Shot                             & GPT-4o                  & \textbf{82.86}             & \textbf{79.52}                     & \textbf{50.63}              & \textbf{39.73}            & \textbf{56.46}             & \textbf{49.86}            \\
\hline
\end{tabular}
}
\label{sfer}
\end{table*}

\begin{table}
	\centering
 \caption{Comparisons of 3D visual grounding on  Sr3D~\cite{achlioptas2020referit3d} and Nr3D~\cite{achlioptas2020referit3d}. We evaluate the top-1 accuracy using ground-truth boxes. ``super": supervision method.} 
	%\resizebox{8cm}{
	\scalebox{0.7}{% Please add the following required packages to your document preamble:
% \usepackage{multirow}
% \usepackage{multirow}\
\small
	\setlength{\tabcolsep}{0.1pt}
%\begin{tabular}{lcccccccccccc}
\begin{tabular}{lccccccccccc}
\hline
\multicolumn{1}{c}{\multirow{2}{*}{\textbf{Method}}} & \multirow{2}{*}{\textbf{Super}} & \multicolumn{5}{c}{\textbf{Nr3d}}                                                                                                    & \multicolumn{5}{c}{\textbf{Sr3d}}                                                                                \\ \cline{3-12} 
\multicolumn{1}{c}{}                                 &                                       & \textbf{Overall}         & \textbf{Easy}            & \textbf{Hard}            & \textbf{V-Dep.}          & \textbf{V-Indep}         & \textbf{Overall}     & \textbf{Easy}        & \textbf{Hard}        & \textbf{V-Dep.}      & \textbf{V-Indep}     \\ \hline
InstanceRefer~\cite{yuan2021instancerefer5}                                        & Fully                                 & 38.8                     & 46.0                     & 31.8                     & 34.5                     & 41.9                     & 48.0                 & 51.1                 & 40.5                 & 45.8                 & 48.1                 \\
FFL-3DOG~\cite{feng2021free}                                             & Fully                                 & 41.7                     & 48.2                     & 35.0                     & 37.1                     & 44.7                     & -                    & -                    & -                    & -                    & -                    \\
LAR~\cite{bakr2022looklar}                                                  & Fully                                 & 48.9                     & 58.4                     & 42.3                     & 47.4                     & 52.1                     & 59.4                 & 63.0                 & 51.2                 & 50.0                 & 59.1                 \\
SAT~\cite{yang2021sat}                                                  & Fully                                 & 56.5                     & 64.9                     & 48.4                     & 54.4                     & 57.6                     & 57.9                 & 61.2                 & 50.0                 & 49.2                 & 58.3                 \\
3D-SPS~\cite{luo20223dsdsps}                                               & Fully                                 & 51.5                     & 58.1                     & 45.1                     & 48.0                     & 53.2                     & 62.6                 & 56.2                 & 65.4                 & 49.2                 & 63.2                 \\
3DJCG~\cite{cai20223djcg}                                                & Fully                                 &    52.4                      &   59.3                       &  47.6                        &46.8                          &   55.9                       &     58.5                 &   64.7                   &       51.8               &  52.0                    &    65.6                  \\
BUTD-DETR~\cite{jain2022bottom}                                            & Fully                                 & 54.6                     & 60.7                     & 48.4                     & 46.0                     & 58.0                     & 67.0                 & 68.6                 & 63.2                 & 53.0                 & 67.6                 \\
MVT~\cite{huang2022multimvt}                                                  & Fully                                 & 59.5                     & 67.4                     & 52.7                     & 59.1                     & 60.3                     & 64.5                 & 66.9                 & 58.8                 & 58.4                 & 58.4                 \\
ViL3DRel~\cite{chen2022languageViL3DRel}                                             & Fully                                 & 64.4                     & 70.2                     & 57.4                     & 62.0                     & 64.5                     & 72.8                 & 74.9                 & 67.9                 & 63.8                 & 73.2                 \\
EDA~\cite{wu2023eda}                                                  & Fully                                 & 52.1                     & 58.2                     & 46.1                     & 50.2                     & 53.1                     & 68.1                 & 70.3                 & 62.9                 & 54.1                 & 68.7                 \\
%3D-VisTA (scratch)~\cite{yang2023llmgrounderopenvocabulary3dvisual}                                   & Fully                                 & 57.5                     & 65.9                     & 49.4                     & 53.7                     & 59.4                     & 69.6                 & 72.1                 & 63.6                 & 57.9                 & 70.1                 \\
3D-VisTA~\cite{yang2023llmgrounderopenvocabulary3dvisual}                                              & Fully                                 & 64.2                     & 72.1                     & 56.7                     & 61.5                     & 65.1                     & 76.4                 & 78.8                 & 71.3                 & 58.9                 & 77.3                 \\
Scene-Verse~\cite{jia2024sceneverse}                                          & Fully                                 & 64.9                     & 72.5                     & 57.8                     & 56.9                     & 67.9                     & 77.5                 & 80.1                 & 71.6                 & 62.8                 & 78.2                 \\ 
TSP3D                                & Fully       & 48.7&- &- &- &- &57.1& -&- &- & -\\
MPEC & Fully & 66.7&- &- &- &- & 80.0& -&- &- & -\\
\hline
WS-3DVG~\cite{gadre2023cows10}                                              & Weakly                                & {27.3} & {18.0} & {21.6} & {22.9} & {22.5} & -                    & -                    & -                    & -                    & -                    \\ \hline
ZSVG3D~\cite{yuan2024visual}                                          & Zero-Shot                             & {46.5} & {31.7} & {36.8} & {40.0} & {39.0} & -                    & -                    & -                    & -                    & -                    \\
% ConceptGraph                                         & Zero-Shot                             & \multicolumn{1}{l}{}     & \multicolumn{1}{l}{}     & \multicolumn{1}{l}{}     & \multicolumn{1}{l}{}     & \multicolumn{1}{l}{}     & \multicolumn{1}{l}{} & \multicolumn{1}{l}{} & \multicolumn{1}{l}{} & \multicolumn{1}{l}{} & \multicolumn{1}{l}{} \\
% BBQ                                                  & Zero-Shot                             & \multicolumn{1}{l}{}     & \multicolumn{1}{l}{}     & \multicolumn{1}{l}{}     & \multicolumn{1}{l}{}     & \multicolumn{1}{l}{}     & \multicolumn{1}{l}{} & \multicolumn{1}{l}{} & \multicolumn{1}{l}{} & \multicolumn{1}{l}{} & \multicolumn{1}{l}{} \\
% Open3DSG                                             & Zero-Shot                             & \multicolumn{1}{l}{}     & \multicolumn{1}{l}{}     & \multicolumn{1}{l}{}     & \multicolumn{1}{l}{}     & \multicolumn{1}{l}{}     & \multicolumn{1}{l}{} & \multicolumn{1}{l}{} & \multicolumn{1}{l}{} & \multicolumn{1}{l}{} & \multicolumn{1}{l}{} \\
SeeGround~\cite{li2025seeground}                                           & Zero-Shot                             & {54.5} & {38.3} &{42.3} & {48.2} & 46.1 & -                    & -                    & -                    & -                    & -                    \\ 
ConceptGraph*~\cite{gu2023conceptgraphsopenvocabulary3dscene41} & Zero-Shot &38.2 &39.4 &32.6 &42.1 &38.7 &43.6 &44.3 &41.9 &38.4 &49.7 \\
BBQ*~\cite{linok2024barequeriesopenvocabularyobject42} & Zero-Shot &45.0  &47.6 &41.2 &46.3 &45.0&49.9 &53.9 &49.3 &45.6 &50.7 \\
VLM-Grounder~\cite{xu2024vlm}       & Zero-Shot  &48.0 &55.2 &39.5 &45.8 &49.4 & - & - & - & - & - \\ 
SORT3D~\cite{zantout2025sort3d} & Zero-Shot &60.5 &-&-&56.6& 62.3 & - & - & - & - & -\\
CSVG~\cite{yuan2024solving} & Zero-Shot  &59.2 &-&-&53.0 &62.5& - & - & - & - & -\\
%ChainSP~\cite{shi2025chain}                                           & Zero-Shot                             & {60.8} & 68.3 &53.3& 50.7& 46.1                & 91.4               & 91.8                  & 90.5                   &54.3            & 92.8      \\ 
%3DLFVG & Zero-Shot  &&&&&&&&&& \\
\hline
Ours                                                 & Zero-Shot                             &  \textbf{61.8}  &   \textbf{61.4}  & \textbf{57.8} &  \textbf{60.9}  & \textbf{67.1}    & \textbf{70.9} &  \textbf{79.3}  &  \textbf{63.9}   &\textbf{64.1} &\textbf{76.5} \\ \hline
\end{tabular}
}
\label{grounding1}
\end{table}
\textbf{Replica}~\cite{replica19arxiv} is a dataset of 18 realistic 3D indoor scene reconstructions, covering rooms to buildings. We selected 8 scene data samples (room0, room1, room2, office0, office1, office2, office3, office4) with their annotations.

\textbf{ScanNet~\cite{dai2017scannet}} is an instance-level indoor RGB-D dataset containing both 2D and 3D data. We selected 8 scene samples, which are 0011, 0030, 0046, 0086, 0222, 0378, 0389, 0435.
%To evaluate the model's ability to answer complex semantic queries, we followed the same evaluation setup as BBQ~\cite{linok2024barequeriesopenvocabularyobject42}, testing with Sr3D~\cite{achlioptas2020referit3d} and Nr3D~\cite{achlioptas2020referit3d} from the ScanNet annotations, and ScanRefer~\cite{chen2020scanrefer} datasets.

\textbf{3DSSG} dataset~\cite{3DSSG202036} offers annotated 3D semantic scene graphs. Adopting the RIO27 annotation, we evaluate 27 classes and 16 relationship classes and adhere to the experimental protocol of EdgeGCN~\cite{zhang2021exploitingedgeorientedreasoning3dsgg} for the fair comparison, dividing the dataset into train/val/test sets with 1084/113/113 scenes. All the camera viewpoints follow the original dataset settings.

%More details of datasets are in Appendix C.
%Metrics include Acc@0.25 IoU and Acc@0.5 IoU, reported for both unique and multiple object categories.

\subsubsection{Performance Metric} 
For visual grounding on the Sr3D~\cite{achlioptas2020referit3d} and Nr3D~\cite{achlioptas2020referit3d}, we follow the ReferIt3D~\cite{achlioptas2020referit3d} protocol by using ground-truth object masks and measuring grounding accuracy, whether the model correctly identifies the target object among the ground-truth proposals. Additionally, to ensure a fair comparison with related works~\cite{gu2023conceptgraphsopenvocabulary3dscene41,linok2024barequeriesopenvocabularyobject42,koch2024open3dsg}, we also report Acc@0.1 IoU and Acc@0.25 IoU for the ``easy", ``hard", ``view-dep." and ``view-indep." cases.
%For visual grounding on the Sr3D~\cite{achlioptas2020referit3d} and Nr3D~\cite{achlioptas2020referit3d} datasets, we adopt the strategy from ReferIt3D~\cite{achlioptas2020referit3d}, utilizing the ground-truth object masks and evaluating grounding performance based on accuracy specifically, whether the model successfully identifies the target object from the set of ground-truth proposals. Besides, to enable a fair comparison with several closely related works~\cite{gu2023conceptgraphsopenvocabulary3dscene41,linok2024barequeriesopenvocabularyobject42,koch2024open3dsg}, we also report Acc@0.1 IoU and Acc@0.25 IoU for the ``easy", ``hard", ``view-dep." and ``view-indep." cases.
%Besides, we compute the  Acc@0.1 IoU and Acc@0.25 IoU for ``easy", ``hard", ``view-dep." and ``view-indep." cases. 
For ScanRefer~\cite{chen2020scanrefer}, we calculate the metrics including Acc@0.25 IoU and Acc@0.5 IoU, reported for both unique, multiple, and overall categories. 
``Unique" which refers to scenes containing only one target object, ``Multiple" which includes scenes with distractor objects from the same class, and ``Overall" which represents the aggregated results across all scene categories. 
For Replica~\cite{replica19arxiv} and ScanNet~\cite{dai2017scannet}, We compute the metrics of mAcc, mIoU, and fmIoU. For 3DSSG~\cite{3DSSG202036}, we adopt the widely used top-k recall metric (R@k) for scene graph evaluation, assessing objects, predicates, and relationships separately. 
For assessment, as shown in Table~\ref{graphcomparison}, Recall@5 and Recall@10 are used for object classification, Recall@3 and Recall@5 for predicate classification, and Recall@50 and Recall@100 for relationship classification.
For out-of-word queries, we validate results using manually annotated ground truth, which will be publicly available. For 5 tested datasets, we all follow the original queries and annotations.

\begin{figure*}
\centering
\includegraphics[width=0.85\linewidth]{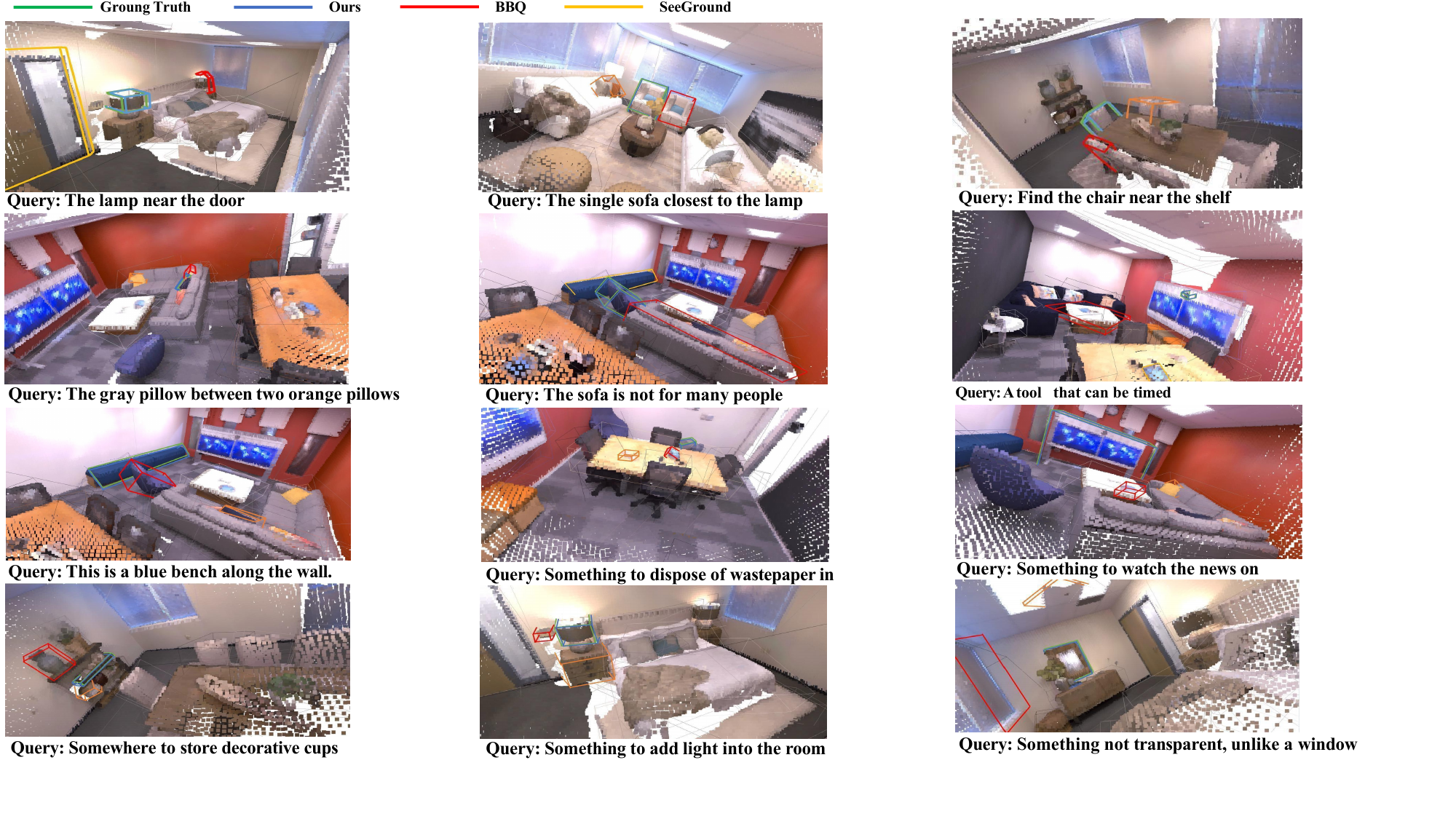}
\caption{Comparison of 3D object visual grounding task with free-form query. The ground truth box is in green.}
\label{grouding22}
\end{figure*}

\begin{table*}
	\centering
 \caption{Comparisons of 3D visual grounding on  Sr3D~\cite{achlioptas2020referit3d} and Nr3D~\cite{achlioptas2020referit3d} datasets. The Accuracy (A) at 0.1 and 0.25 IoU thresholds is presented separately for 5 categories.} 
	%\resizebox{8cm}{
	\scalebox{0.75}{% Please add the following required packages to your document preamble:
% \usepackage{multirow}
% \usepackage{multirow}\
\small
	\setlength{\tabcolsep}{0.1pt}
%\begin{tabular}{lcccccccccccc}
\begin{tabular}{l|cccccccccccccccccccc}
\hline
    \multicolumn{1}{c|}{\multirow{3}{*}{Methods}} & \multicolumn{10}{c|}{Sr3D}                                                                                                                                                        & \multicolumn{10}{c}{Nr3D}                                                                                                                                       \\ \cline{2-21} 
\multicolumn{1}{c|}{}                         & \multicolumn{2}{c}{Overall}   & \multicolumn{2}{c}{Easy}      & \multicolumn{2}{c}{Hard}      & \multicolumn{2}{c}{View Dep.} & \multicolumn{2}{c|}{View Indep.}                  & \multicolumn{2}{c}{Overall}   & \multicolumn{2}{c}{Easy}      & \multicolumn{2}{c}{Hard}      & \multicolumn{2}{c}{View Dep.} & \multicolumn{2}{c}{View Indep.} \\ \cline{2-21} 
\multicolumn{1}{c|}{}                         & { A@0.1}   & {A@0.25}  & {A@0.1}   & { A@0.25}  & { A@0.1}   & {A@0.25}  & { A@0.1}   & {A@0.25}  & { A@0.1}   & \multicolumn{1}{c|}{{ A@0.25}} & { A@0.1}   & { A@0.25}  & { A@0.1}   & { A@0.25}  & {A@0.1}   & { A@0.25}  & { A@0.1}   & {A@0.25}  & { A@0.1}    & { A@0.25}   \\ \hline
OpenFusion~\cite{zhang2024towards}                                    & 12.6          &{2.4}           & 14.0          & 2.4           & 1.3           & 1.3           & 3.8           & 2.5           & 13.7          & \multicolumn{1}{c|}{2.4}                               & 10.7          & 1.4           & 12.9          & 1.4           & 5.1           & 1.5           & 8.5           & 0.0           & 11.4           & 1.9            \\
ConceptGraph~\cite{gu2023conceptgraphsopenvocabulary3dscene41}                                  & 13.3          & 6.2           & 13.0          & 6.8           & 16.0          & 1.3           & 15.2          & 5.1           & 13.1          & \multicolumn{1}{c|}{6.4}                               & 16.0          & 7.2           & 18.7          & 9.2           & 9.1           & 2.0           & 12.7          & 4.2           & 17.0           & 8.1            \\
BBQ~\cite{linok2024barequeriesopenvocabularyobject42}                                           & 34.2          & 22.7          & 34.3          & 22.7          & 33.3          & 22.7          & 32.9          & 20.3          & 34.4          & \multicolumn{1}{c|}{23.0}                              & 28.3          & 19.0          & 30.5          & 21.3          & 22.8          & 13.2          & 23.6          & 18.2          & 29.8           & 19.3           \\ 
Open3DSG~\cite{koch2024open3dsg}                                    &    37.3                                                &  {25.8} &36.5& 24.8& 36.1& 25.9&36.1&23.2 &36.6&\multicolumn{1}{c|}{25.3}&31.4 &22.5&32.3&24.3&25.2&16.8&24.3&21.2&33.5&22.8 \\ 
3DLFVG~\cite{zhang2024towards} &- & 18.3   & -& 15.1 & -  &  19.0  &- & 18.3   &- & \multicolumn{1}{c|}{22.4}    &- & 19.3  & - & 21.0   & - & 15.2 & -  & 11.2&-&  19.1 \\
\hline
\textbf{Ours}                                 & \textbf{61.1} & \textbf{46.3} & \textbf{60.8} & \textbf{46.4} & \textbf{57.8} & \textbf{54.9} & \textbf{59.3} & \textbf{51.2} & \textbf{60.9} & \multicolumn{1}{c|}{\textbf{52.6}}                   & \textbf{51.3} & \textbf{43.5} & \textbf{58.2} & \textbf{45.9} & \textbf{47.4} & \textbf{39.8} & \textbf{50.3} & \textbf{47.6} & \textbf{54.3}  & \textbf{38.9}  \\ \hline
\end{tabular}
}
\label{grounding}
\end{table*}

\subsubsection{Implementation Details}
We conduct experiments on the NVIDIA 3090 GPU using PyTorch. 
We adopt GPT-4o~\cite{achiam2023gpt} as the LLM, LLaVa-7B-v1.6~\cite{liu2024visualLLAVA} as the LVLM. For 2D objects and encoding, we use Grounded-SAM~\cite{ren2024grounded} for 2d mask segmentation and employ the CLIP  ViT-L/14 encoder~\cite{radford2021learning} as the visual feature extractor. 
%With thresholds set at ${\tau _{iou}} = 0.9$ and ${\tau _{sim}} = 0.9$, 
We select the top-5 view masks with the highest projected point IoU for each superpoint. Following ConceptGraph~\cite{gu2023conceptgraphsopenvocabulary3dscene41}, for each object, we select relevant image crops from the Top-10 best views and pass them to LLM to generate captions. %More settings and ablation studies are in Appendix C and E.
For superpoint merging, we employ consistent thresholds of $\tau_{iou}$ = 0.9 and $\tau_{sim}$= 0.9 across all experiments. 
For each superpoint, we select the top-5 view masks with the highest IoU relative to the projected points. Following ConceptGraph~\cite{gu2023conceptgraphsopenvocabulary3dscene41}, we set the voxel size and nearest neighbor threshold to 2.5 cm, and use an association threshold of 1.1.
%For each superpoint, we select the Top-5 view masks that exhibit the highest IoU with the projected points. Similar to ConceptGraph~\cite{gu2023conceptgraphsopenvocabulary3dscene41}, we set the voxel size for point cloud downsampling and the nearest neighbor threshold to 2.5 cm. Additionally, an association threshold of 1.1 is employed.

\subsection{Experiment Results}
\subsubsection{3D Object Grounding}
We conducted 3D visual grounding comparisons on the Nr3D~\cite{achlioptas2020referit3d}, Sr3D~\cite{achlioptas2020referit3d}, and ScanRefer~\cite{chen2020scanrefer} datasets. 
As shown in Table~\ref{sfer}, we conducted comprehensive experimental comparisons on the ScanRefer~\cite{chen2020scanrefer} benchmark, evaluating a wide range of models across different learning paradigms. These include state-of-the-art fully supervised approaches~\cite{yuan2021instancerefer5,zhao20213dvg1,jain2022bottom,yang2023llmgrounderopenvocabulary3dvisual,cai20223djcg,wu2021scenegraphfusionincremental3dscene39,koch2023sgrec3dselfsupervised3dscene,jia2024sceneverse}, weakly supervised methods~\cite{gadre2023cows10}, fine-tuned models~\cite{peng2023openscene7,huang2024chat,zhu2024scanreasonreground,yu2025inst3d} refer to methods that are adapted to specific tasks after fine-tuning, and zero-shot approaches~\cite{koch2024open3dsg,kerr2023lerf,peng2023openscene7,gu2023conceptgraphsopenvocabulary3dscene41,linok2024barequeriesopenvocabularyobject42,yang2023llmgrounderopenvocabulary3dvisual,li2025seeground} refer to methods that directly use LLMs without fine-tuning.
%As shown in Table~\ref{sfer}, we conducted extensive experimental comparisons on ScanRefer~\cite{chen2020scanrefer}, evaluating models trained with advanced fully-supervised learning models~\cite{yuan2021instancerefer5,zhao20213dvg1,jain2022bottom,yang2023llmgrounderopenvocabulary3dvisual,cai20223djcg,wu2021scenegraphfusionincremental3dscene39,koch2023sgrec3dselfsupervised3dscene,cai20223djcg,jia2024sceneverse}, weakly-supervised learning model~\cite{gadre2023cows10}, and zero-shot learning models~\cite{koch2024open3dsg,kerr2023lerf,peng2023openscene7,gu2023conceptgraphsopenvocabulary3dscene41,linok2024barequeriesopenvocabularyobject42,yang2023llmgrounderopenvocabulary3dvisual,li2025seeground}. 
Our model, without requiring any training, achieved best results, fully demonstrating its superiority. %Specifically, although we did not outperform the fully-supervised models, such as Scene-Verse~\cite{jia2024sceneverse} and MCLN~\cite{mei2024vocabulary}, which require substantial training resources and time on 3D datasets, our model, being training-free, achieved nearly equivalent results. This highlights both the superiority and efficiency of our model. 
While our training-free model does not surpass fully-supervised or fine-tuned approaches like Scene-Verse~\cite{jia2024sceneverse} and Inst3D-LLM~\cite{yu2025inst3d}, which demand extensive training on 3D data and LLMs, it achieves comparable performance without any training cost, underscoring its efficiency and effectiveness.
Furthermore, compared to the zero-shot models, our model achieved best results across all categories with a clear advantage. To further substantiate our findings, we also performed ablation experiments on different LLM agents, which further demonstrates that our model consistently yields optimal results across various LLMs.

%To further substantiate our findings, we also performed ablation experiments on different LLM agents, as detailed in Table~\ref{llm}, which further demonstrates that our model consistently yields optimal results across various LLMs .

\begin{figure*}
\centering
\includegraphics[width=0.85\linewidth]{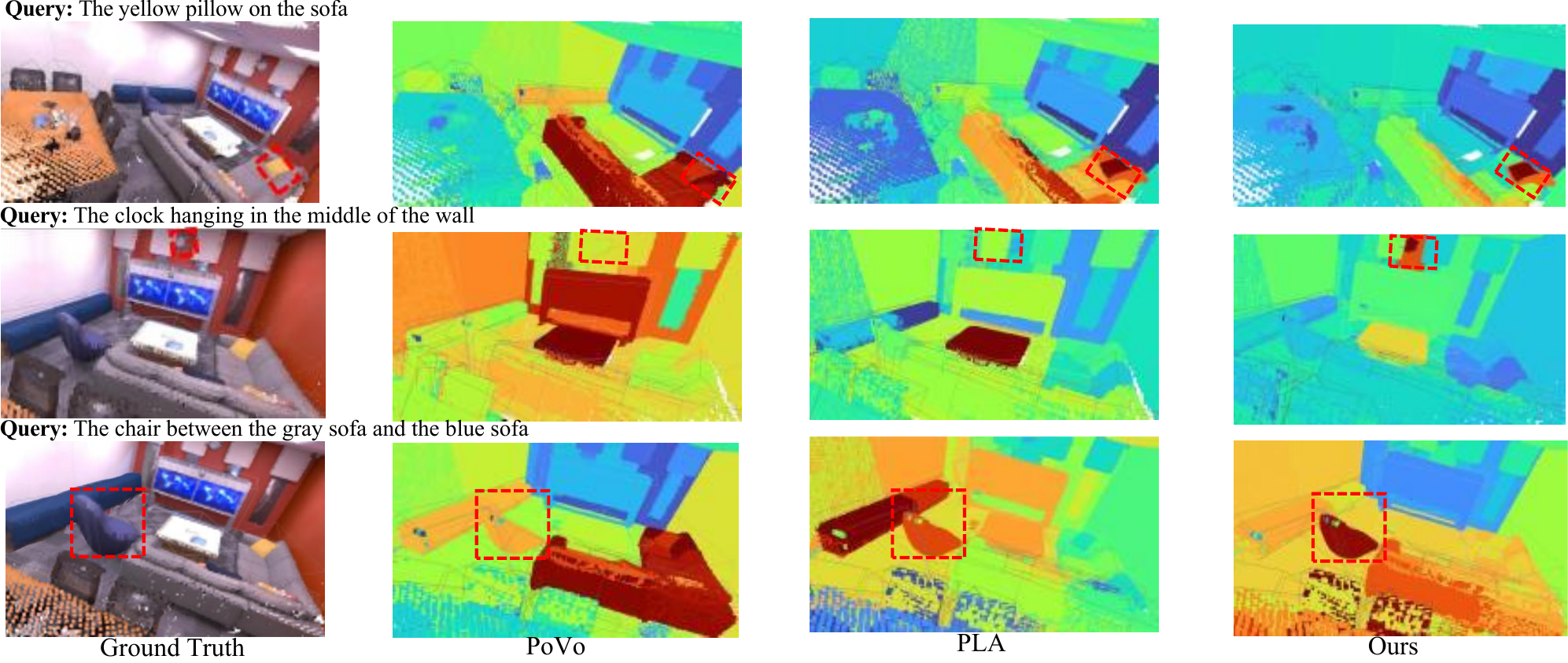}
\caption{Comparison semantic segmentation on the Replica dataset. The semantic map highlights the regions most relevant to the query's semantic features, with deeper colors indicating higher relevance, where red represents the most relevant semantics.}
%\caption{Comparison of three types of complex semantic query on the Replica dataset.}
\label{complexsemantic}
\end{figure*}

\begin{table}
	\centering
 \caption{Comparisons of 3D semantic segmentation task between our model and SOTA methods on Replica and ScanNet datasets.} 
	%\resizebox{8cm}{
	\scalebox{0.8}{% Please add the following required packages to your document preamble:
% \usepackage{multirow}
% \usepackage{multirow}\
\small
	\setlength{\tabcolsep}{1pt}
%\begin{tabular}{lcccccccccccc}
\begin{tabular}{lcllllll}
\hline
\multicolumn{1}{c|}{\multirow{2}{*}{\textbf{Method}}} & \multicolumn{1}{c|}{\multirow{2}{*}{\textbf{Type}}} & \multicolumn{3}{c|}{\textbf{Replica}}                                                                & \multicolumn{3}{c}{\textbf{ScanNet}}                                                      \\ \cline{3-8} 
\multicolumn{1}{c|}{}                                 & \multicolumn{1}{c|}{}                      & \multicolumn{1}{c|}{\textbf{mAcc↑}} & \multicolumn{1}{c|}{{ \textbf{mIoU↑}}} & \multicolumn{1}{c|}{\textbf{fmIoU↑}} & \multicolumn{1}{c|}{{ \textbf{mAcc↑}}} & \textbf{mIoU↑}                 & \textbf{fmIoU↑}                \\ \hline
LSeg~\cite{li2022language}& \multirow{3}{*}{Privileged}                & 0.33                       & -                                & 0.40                        & \multicolumn{1}{c}{-}            & \multicolumn{1}{c}{-} & \multicolumn{1}{c}{-} \\
OpenSeg~\cite{ghiasi2022scalingopenvocabularyimagesegmentation}                                               &                                            & 0.41                       & -                                & 0.54                        & \multicolumn{1}{c}{-}            & \multicolumn{1}{c}{-} & \multicolumn{1}{c}{-} \\
OpenFusion~\cite{yamazaki2024open}                                            &                                            & 0.41                       & 0.30                              & 0.58                        & 0.67                             & 0.53                  & 0.64                  \\ 
OpenScene~\cite{peng2023openscene7}& &- & - &- &0.69& &0.58 \\
MPEC~\cite{wang2025masked} & &- &- &- &0.79 &- &0.65 \\
\hline
Mask2former~\cite{cheng2021maskformer}                                           & \multirow{5}{*}{Zero-shot}                 & 0.05                       & -                                & 0.01                        & \multicolumn{1}{c}{-}            & \multicolumn{1}{c}{-} & \multicolumn{1}{c}{-} \\
ConceptFusion~\cite{jatavallabhula2023conceptfusion12}                                         &                                            & 0.29                       & 0.11                             & 0.14                        & 0.49                             & 0.26                  & 0.31                  \\
OpenMask3D~\cite{takmaz2023openmask3d}                                            &                                            & -                          & -                                & -                           & 0.34                             & 0.18                  & 0.20                   \\
ConceptGraphs~\cite{gu2023conceptgraphsopenvocabulary3dscene41}                                         &                                            & 0.36                       & 0.18                             & 0.15                        & 0.52                             & 0.26                  & 0.29                  \\
HOV-SG~\cite{werby2024hierarchical}                                   &                                            & 0.29                    & 0.23                            & -                 & 0.44                             & 0.22                  & 0.29                  \\
BBQ-CLIP~\cite{linok2024barequeriesopenvocabularyobject42}                                              &                                            & 0.38                       & 0.27                             & 0.48                        & 0.56                             & 0.34                  & 0.36                  \\ 
Open3DSG~\cite{koch2024open3dsg} & 
& 0.35                       & 0.31                             & 0.45                      & 0.52                             & 0.36                  & 0.31         
\\ 
PoVo~\cite{mei2024vocabulary} & 
& 0.34                       &   0.35                       &           0.52          &     0.34                       &           0.39     & 0.36    
\\ 
PLA~\cite{ding2023pla9} & 
& -                    &   -                    &           -         &     0.41                      &          -    & 0.19
\\ 
\hline
\textbf{Ours}                                                  & \multicolumn{1}{l}{Zero-shot}              & \textbf{0.49}                       & \textbf{0.42}                             & \textbf{0.66}                        & \textbf{0.82}                             & \textbf{0.57}                  & \textbf{0.69}                  \\ \hline
\end{tabular}}
\label{seg}
\end{table}
Besides, for Sr3D and Nr3D datasets, we evalute the top-1 using groud-truth boxes in Table~\ref{grounding1} and the accuray at 0.1 and 0.25 IOU threshold for 5 categories in Table~\ref{grounding}.
%under identical experimental settings with the same LVLM model across all methods, 
As shown in Table~\ref{grounding1}, we validated the top-1 performance using ground-truth boxes across fully-supervised models~\cite{yuan2021instancerefer5,cai20223djcg,yang2023llmgrounderopenvocabulary3dvisual}, weakly-supervised model~\cite{gadre2023cows10}, and zero-shot models~\cite{li2025seeground,yuan2024visual}. Our model achieved best results across 5 different metrics, demonstrating its superior performance.
As shown in Table~\ref{grounding}, our model also consistently outperformed all SOTA works~\cite{yamazaki2024open, gu2023conceptgraphsopenvocabulary3dscene41,linok2024barequeriesopenvocabularyobject42} across 4 cases of  Sr3D and Nr3D datasets. 
%Compared with OpenFusion~\cite{yamazaki2024open}, which depends on CLIP for object grounding, our model effectively maps free-form objects and relations, addressing CLIP's bag-of-words limitations and semantic inconsistencies. 
Compared to ConceptGraph~\cite{gu2023conceptgraphsopenvocabulary3dscene41}, BBQ~\cite{linok2024barequeriesopenvocabularyobject42} and Open3DSG~\cite{koch2024open3dsg}, which also utilize LLMs and graph representations for reasoning, our model shows significant advantages, validating the semantic aligned features in reasoning with free-form queries.  
%Furthermore, we compare our model with CLIP-based~\cite{kerr2023lerf,peng2023openscene7,zhang2024towards} and LLM-based~\cite{yang2023llmgrounderopenvocabulary3dvisual,linok2024barequeriesopenvocabularyobject42} SOTA methods. As shown in Table~\ref{sfer}, our approach achieves significant improvements over LLM-based methods and demonstrates superior advantages over CLIP-based ones. 
%Besides, Unlike OpenFusion~\cite{yamazaki2024open}, which relies on CLIP encoders for object grounding, our semantic consistent 3D scene graph representation effectively interpret complex grounding tasks while overcoming CLIP's inherent bag-of-words limitations and semantic inconsistencies. 
Fig.\ref{grouding22} shows the quantitative comparison. SeeGround~\cite{li2025seeground} fails to capture object relationships like ``near", while BBQ~\cite{linok2024barequeriesopenvocabularyobject42} struggles with semantic label like ``single sofa", hindering accurate grounding. In contrast, our model precisely grounds objects with correct semantic labels and understands both scene and object-level spatial relationships. %We present more comparisons in Appendix D.
%As illustrated in rows 3 to 4 of Fig.~\ref{grouding22}, we further present a comparison of our model against BBQ~\cite{linok2024barequeriesopenvocabularyobject42} and SeeGround~\cite{li2025seeground} in 3D visual grounding with complex free-form semantic queries. The results demonstrate that our model consistently identifies the correct target objects under various complex semantic queries, whereas others struggle to comprehend and resolve such intricate semantics.

\begin{table}
	\centering
 \caption{Comparisons of 3D scene graph generation in object, predicate, and relationship prediction on 3DSSG~\cite{3DSSG202036} dataset.} 
	%\resizebox{8cm}{
	\scalebox{0.8}{% Please add the following required packages to your document preamble:
% \usepackage{multirow}
% \usepackage{multirow}\
\small
	\setlength{\tabcolsep}{4pt}
%\begin{tabular}{lcccccccccccc}
\begin{tabular}{lcccccc}
\hline
\multicolumn{1}{c}{\multirow{2}{*}{\textbf{Method}}} & \multicolumn{2}{c}{\textbf{Object}}    & \multicolumn{2}{c}{\textbf{Predicate}} & \multicolumn{2}{c}{\textbf{Relationship}} \\ \cline{2-7} 
\multicolumn{1}{c}{}                        & \textbf{R@5}           & \textbf{R@10}          & \textbf{R@5}           & \textbf{R@10}          & \textbf{R@50}             & \textbf{R@100}           \\ \hline
\emph{\textbf{Fully-supervised}} \\
3DSSG~\cite{3DSSG202036}    & 0.68        & 0.78        & 0.89        & 0.93          & 0.40          &0.66           \\ 
% VL-SAT~\cite{wang2023vl}                                    & 0.57          & 0.68          & 0.63          & 0.70          & 0.64            & 0.66           \\ 
SGFN~\cite{wu2021scenegraphfusionincremental3dscene39}                                    &0.70  & 0.80              & 0.97          & 0.99          & 0.85     &0.87    \\
SGRec3D~\cite{koch2023sgrec3dselfsupervised3dscene}                              &0.80  & 0.87              & 0.97          & 0.99          & 0.89     &0.91    \\
SGPN~\cite{3DSSG202036}  & 0.68        & 0.78        & 0.89        & 0.93         & 0.40            & 0.66           \\
  VL-SAT~\cite{wang2023vl}    & \textbf{0.78}         & \textbf{0.86}          & \textbf{0.98}         & \textbf{0.99}         & \textbf{0.90}            & \textbf{0.93}           \\  
\hline
\textbf{\emph{Zero-shot open-vocabulary}} \\
OpenSeg~\cite{ghiasi2022scalingopenvocabularyimagesegmentation}                                     & 0.38          & 0.45          & 0.10          & 0.23          & 0.05            & 0.07           \\
ConceptGraphs~\cite{gu2023conceptgraphsopenvocabulary3dscene41}                               & 0.41          & 0.48          & 0.39          & 0.47          & 0.32            & 0.28           \\
BBQ~\cite{linok2024barequeriesopenvocabularyobject42}                                         & 0.43          & 0.54          & 0.46          & 0.54          & 0.41            & 0.42           \\
Open3DSG~\cite{koch2024open3dsg}                                   & 0.57          & 0.68          & 0.63          & 0.70          & 0.64            & 0.66                 \\
\textbf{Ours}                               & \textbf{0.69} & \textbf{0.78} & \textbf{0.90} & \textbf{0.94} & \textbf{0.82}   & \textbf{0.88}  \\ \hline
\end{tabular}}
\label{graphcomparison}
\end{table}

\subsubsection{Complex Queries}
To evaluate our model's capability for complex semantic queries, we compare the ``hard" case on Sr3D~\cite{achlioptas2020referit3d} and Nr3D~\cite{achlioptas2020referit3d} datasets, and ``Multiple" case on ScanRefer~\cite{chen2020scanrefer}. 
As shown in Tables~\ref{grounding} and \ref{sfer}, our model exhibits significant advantages in handling all complex semantic queries and multi-object queries. 
%Compared to BBQ~\cite{linok2024barequeriesopenvocabularyobject42}, our model achieves improvements of 22.2 and 16.6 in A@0.25 accuracy on the ``hard" cases of the Nr3D and Sr3D datasets, respectively. 
This validates that our approach can more effectively comprehend complex semantic queries, leveraging 3D semantically consistent scene graphs. 
As illustrated in rows 3-4 of Fig.~\ref{grouding22}, we further present a comparison of our model against BBQ~\cite{linok2024barequeriesopenvocabularyobject42} and SeeGround~\cite{li2025seeground} in 3D visual grounding with complex free-form semantic queries. The results demonstrate that our model consistently identifies the correct target objects under various complex semantic queries, whereas others struggle to comprehend and resolve such intricate semantics.

\subsubsection{3D Semantic Segmentation}
As shown in Table~\ref{seg} and Fig.~\ref{complexsemantic}, we evaluate on 3D semantic segmentation task on Replica~\cite{replica19arxiv} and ScanNet~\cite{dai2017scannet} datasets. Following ConceptGraph~\cite{gu2023conceptgraphsopenvocabulary3dscene41}, we matched object nodes' fused features to CLIP text embeddings of \emph{``an image of class''}, then assigned points to their semantic categories via similarity scores. 
We compare our model against SOTA zero-shot 3D open-vocabulary segmentation methods~\cite{jatavallabhula2023conceptfusion12,cheng2021maskformer,takmaz2023openmask3d,gu2023conceptgraphsopenvocabulary3dscene41,linok2024barequeriesopenvocabularyobject42} and privileged approaches leveraging pre-trained datasets~\cite{ghiasi2022scalingopenvocabularyimagesegmentation,zhang2024towards,li2022language}, where our method consistently achieves notable gains. Compared to BBQ~\cite{linok2024barequeriesopenvocabularyobject42} and Open3DSG~\cite{koch2024open3dsg}, our model delivers superior results on the ScanNet benchmark~\cite{dai2017scannet}. Furthermore, our zero-shot approach surpasses OpenFusion~\cite{yamazaki2024open}, a supervised model fine-tuned for semantic segmentation, highlighting the strength of our training-free framework.
%We compare our model with SOTA zero-shot 3D open-vocabulary segmentation ~\cite{jatavallabhula2023conceptfusion12,cheng2021maskformer,takmaz2023openmask3d,gu2023conceptgraphsopenvocabulary3dscene41,linok2024barequeriesopenvocabularyobject42} and privileged methods~\cite{ghiasi2022scalingopenvocabularyimagesegmentation,zhang2024towards,li2022language} using pre-trained segmentation models, where ours achieves notable improvements.Compared to BBQ~\cite{linok2024barequeriesopenvocabularyobject42} and Open3DSG~\cite{koch2024open3dsg}, our model shows superior performance on the ScanNet~\cite{dai2017scannet}. Besides, our zero-shot approach outperformed OpenFusion~\cite{yamazaki2024open}, a supervised model tailored for semantic segmentation, showing the superior capability of our model without any training priors.
\begin{table}[!htp]
	\centering
 \caption{Ablation study. Graph: 3D scene graph, SA: Semantic Alignment. Reasoning: LLM-based  Reasoning. We present the overall accuracy. } 
	\scalebox{0.75}{
\small
	\setlength{\tabcolsep}{6pt}
\begin{tabular}{ccccccc}
\hline
\multicolumn{3}{c}{\textbf{Method}} & \multicolumn{2}{c}{\textbf{Sr3D}} & \multicolumn{2}{c}{\textbf{Nr3D}} \\ \hline
\textbf{Graph}  & \textbf{SA}  & \textbf{Reasoning}  & \textbf{Acc@0.1}    & \textbf{Acc@0.25}    & \textbf{Acc@0.1}    & \textbf{Acc@0.25}    \\ \hline
×      & ×    & ×          & 13.3       & 6.2         & 16.0       & 7.2         \\
\checkmark      & ×    & ×          & 33.7       & 21.2        & 29.7       & 21.1        \\
\checkmark        & \checkmark      & ×          & 43.3       & 32.2        & 38.9       & 30.3        \\
\checkmark        & \checkmark     & \checkmark           & 61.1       & 46.3        & 51.3       & 43.5        \\ \hline
\end{tabular}
}
\label{ablation}
\end{table}
\begin{figure}[h]
\centering
\includegraphics[width=1\linewidth]{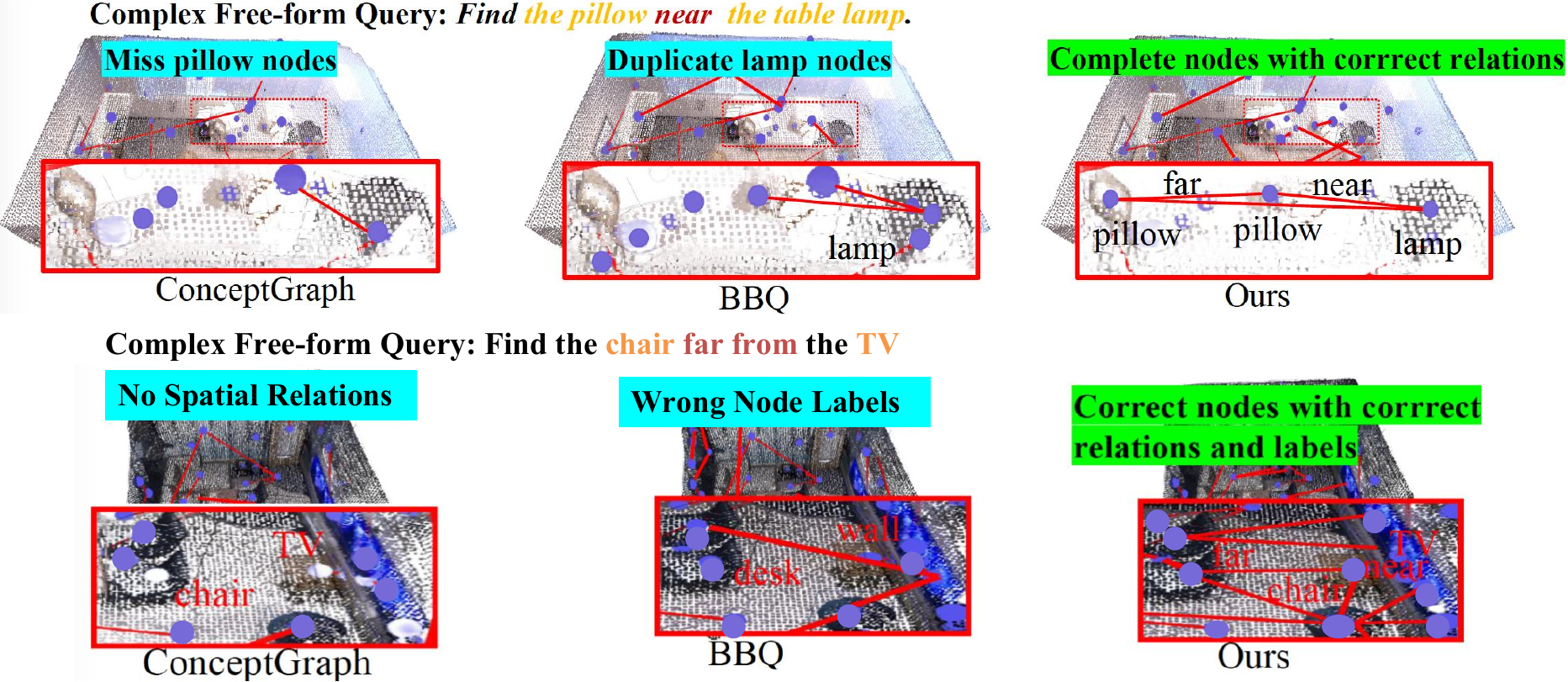}
\caption{Comparison of our semantic consistent scene graph with other scene graphs of the ConceptGraph~\cite{gu2023conceptgraphsopenvocabulary3dscene41} and BBQ~\cite{linok2024barequeriesopenvocabularyobject42}.}
\label{graph}
\end{figure}
In Fig.~\ref{complexsemantic}, following ConceptGraph~\cite{gu2023conceptgraphsopenvocabulary3dscene41}, we compute the similarity between each node's semantic features and the query's CLIP text embedding, with darker map colors (red) indicating higher semantic similarity. Our method {pinpoints key semantic features}, whereas  others fixate on irrelevant cues.
%As illustrated in Fig.~\ref{complexsemantic}, we further present additional comparisons of 3D semantic segmentation with free-form queries on the Replica dataset.  
For various free-form semantic queries, our model accurately segments the corresponding semantic areas, while PoVo~\cite{mei2024vocabulary} and PLA~\cite{ding2023pla9} fail to understand these complex, free-form semantic queries.
\subsubsection{3D Scene Graph} 
%Table~\ref{graphcomparison} shows our 3D scene graph evaluation on 3DSSG~\cite{3DSSG202036}, where we surpass all SOTA models~\cite{ghiasi2022scalingopenvocabularyimagesegmentation,gu2023conceptgraphsopenvocabulary3dscene41,linok2024barequeriesopenvocabularyobject42,koch2024open3dsg} in object, predicate, and relationship prediction. This demonstrates that methods leverage on pre-train datasets, such as OpenSeg~\cite{ghiasi2022scalingopenvocabularyimagesegmentation}, are unsuitable for predicting object nodes and relationships, while models over-rely on LLMs(e.g., ConceptGraph~\cite{gu2023conceptgraphsopenvocabulary3dscene41}, BBQ~\cite{linok2024barequeriesopenvocabularyobject42}) fail on small or thin objects. While Open3DSG~\cite{koch2024open3dsg} can predict open-vocabulary objects, it still faces challenges in free-form relationship prediction. In contrast, our model can precisely predict free-form objects, predicates, and relationships without any training priors.
As shown in Table~\ref{graphcomparison}, we evaluate our 3D scene graph generation on the 3DSSG~\cite{3DSSG202036}. Our method outperforms most fully-supervised models and all zero-shot baselines~\cite{ghiasi2022scalingopenvocabularyimagesegmentation,gu2023conceptgraphsopenvocabulary3dscene41,linok2024barequeriesopenvocabularyobject42,koch2024open3dsg}, while maintaining a training-free and highly efficient pipeline. Although our model does not surpass VL-SAT, which is trained on large-scale vision-language datasets, it achieves competitive performance without requiring any training. Moreover, in contrast to existing methods that over-rely on LLMs (e.g., ConceptGraph~\cite{gu2023conceptgraphsopenvocabulary3dscene41}, BBQ~\cite{linok2024barequeriesopenvocabularyobject42}), our approach demonstrates robustness in predicting small or thin objects, where these methods typically fails.
As shown in Fig~\ref{graph}, it illustrates that our model not only creates a semantically consistent scene graph with complete nodes and correct relations but also assigns the correct semantic labels to each node.

\subsection{Ablation Study}
Ablations are conducted to validate the efficacy of the proposed methods. In the first row of Table 5, we use ConceptGraph~\cite{gu2023conceptgraphsopenvocabulary3dscene41} as a baseline. For the model without reasoning, we apply ConceptGraph's simpler reasoning for inference.

\subsubsection{3D Scene Graph}
The comparison between Rows 1–2 in Table~\ref{ablation} demonstrates that our 3D scene graph significantly improves visual grounding performance on Sr3D and Nr3D. This validates the effectiveness of our scene representation in capturing free-form objects and their relationships. As illustrated in Fig.~\ref{graph}, our model constructs a semantically consistent scene graph with complete nodes and accurate relations.

\begin{table}
	\centering
 \caption{Comparisons of reasoning algorithm on Sr3D and Nr3D.} 
	%\resizebox{8cm}{
	\scalebox{0.75}{% Please add the following required packages to your document preamble:
% \usepackage{multirow}
% \usepackage{multirow}\
\small
	\setlength{\tabcolsep}{1pt}
%\begin{tabular}{lcccccccccccc}
\begin{tabular}{lcccccc}
\hline
\multicolumn{1}{c|}{\multirow{2}{*}{\textbf{Method}}} & \multicolumn{1}{c|}{\multirow{2}{*}{\textbf{Scene Reasoning}}} & \multicolumn{1}{c|}{\multirow{2}{*}{\textbf{Edges}}} & \multicolumn{2}{c|}{\textbf{Sr3D}}                                          & \multicolumn{2}{c}{\textbf{Nr3D}}                      \\ \cline{4-7} 
\multicolumn{1}{c|}{}                                 & \multicolumn{1}{c|}{}                                 & \multicolumn{1}{c|}{}                       & \multicolumn{1}{c|}{{ \textbf{Acc@0.1}}} & \multicolumn{1}{c|}{\textbf{Acc@0.25}} & \multicolumn{1}{c|}{{ \textbf{Acc@0.1}}} & \textbf{Acc@0.25} \\ \hline
BBQ~\cite{linok2024barequeriesopenvocabularyobject42}                                                   & Random                                                & -                                           & 0.02                               & 0.01                          & 0.02                               & 0.01     \\
Ours                                                  & Random                                                & -                                           & 0.21                               & 0.13                          & 0.16                               & 0.14     \\ \hline
ConceptGraphs~\cite{gu2023conceptgraphsopenvocabulary3dscene41}                                         & ConceptGraphs                                         & -                                           & 0.08                               & 0.02                          & 0.07                               & 0.03     \\
ConceptGraphs~\cite{gu2023conceptgraphsopenvocabulary3dscene41}                                         & BBQ                                             & distance                                           & 0.15                               & 0.08                          & 0.12                               & 0.08     \\
ConceptGraphs~\cite{gu2023conceptgraphsopenvocabulary3dscene41}                                         & Ours(two-stage)                                                 & distance                                           & 0.36                               & 0.27                          & 0.33                               & 0.29     \\ \hline
BBQ~\cite{linok2024barequeriesopenvocabularyobject42}                                                   & Deductive                                             & distance                                    & 0.34                              & 0.23                          & 0.28                              & 0.19      \\
BBQ~\cite{linok2024barequeriesopenvocabularyobject42}                                                   & ConceptGraphs                                             & distance                                    &          0.30                  &   0.19                   &    0.25                      & 0.13 \\
BBQ~\cite{linok2024barequeriesopenvocabularyobject42}                                                   & Ours(two-stage)                                              & distance                                    &       0.45                     &   0.28                   &    0.37                     & 0.23  \\ \hline
\textbf{Ours}                                                  & Ours(two-stage)                                             & distance                                    & \textbf{0.61}                               & \textbf{0.46}                          & \textbf{0.51}                               & \textbf{0.44}     \\ \hline
\end{tabular}}
\label{reasoning}
\end{table}
\begin{table}[!htp]
	\centering
      \setlength{\tabcolsep}{1.pt}
\captionsetup{font={small}}
\caption{Comparison of mean and overall computational time.}    \abovedisplayskip=-0.8cm
	\scalebox{0.75}{% Please add the following required packages to your document preamble:
% \usepackage{multirow}
% \usepackage{multirow}\、
\small
\begin{tabular}{lcccccccc}
\hline
\multicolumn{1}{c}{\multirow{2}{*}{Methods}} & \multicolumn{2}{c|}{Replica (average)}                             & \multicolumn{2}{c|}{Nr3d (average)}                                & \multicolumn{2}{c|}{Sr3d (average)}                                & \multicolumn{2}{c}{ScanRefer (avg)}                           \\ \cline{2-9} 
\multicolumn{1}{c}{}                         & \multicolumn{1}{l}{mean(s/it)} & \multicolumn{1}{l|}{overall(min)} & \multicolumn{1}{l}{mean} & \multicolumn{1}{l|}{overall} & \multicolumn{1}{l}{mean} & \multicolumn{1}{l|}{overall} & \multicolumn{1}{c}{mean} & \multicolumn{1}{c}{overall} \\ \hline
ConceptGraph                                 & 3.56                           & \multicolumn{1}{c|}{24.6}                            & 3.98                           & \multicolumn{1}{c|}{25.8}                            & 3.71                           & \multicolumn{1}{c|}{23.4}                             &     4.13                           &        27.6                          \\
BBQ                                          & 1.18                           & \multicolumn{1}{c|}{7.9}                              & 1.24                           & \multicolumn{1}{c|}{8.3}                              & 1.13                           & \multicolumn{1}{c|}{7.6}                              &     1.45                           &      9.3                            \\ \hline
Ours                                         & \textbf{1.03}                           &  \multicolumn{1}{c|}{\textbf{5.6}}                              &  \textbf{1.07}                           &  \multicolumn{1}{c|}{\textbf{6.4}}                              &  \textbf{1.08}                           &  \multicolumn{1}{c|}{\textbf{5.8}}                              &    \textbf{1.16}                            &         \textbf{6.4}                          \\ \hline
\end{tabular}
} 
	\label{time}
\end{table}
\begin{table}[!htp]
	\centering
 \caption{Ablation studies of different LLaVA models.} 
	%\resizebox{8cm}{
	\scalebox{0.75}{% Please add the following required packages to your document preamble:
% \usepackage{multirow}
% \usepackage{multirow}\
\small
	\setlength{\tabcolsep}{3pt}
%\begin{tabular}{lcccccccccccc}
\begin{tabular}{clcccc}
\hline
\multirow{2}{*}{Method}        & \multicolumn{1}{c}{\multirow{2}{*}{LLava type}} & \multicolumn{2}{c}{Sr3D}                    & \multicolumn{2}{c}{Nr3D}                    \\ \cline{3-6} 
                               & \multicolumn{1}{c}{}                            & \multicolumn{1}{l}{Acc@0.1} & Acc@0.25      & \multicolumn{1}{l}{Acc@0.1} & Acc@0.25      \\ \hline
\multirow{3}{*}{ConceptGraphs~\cite{gu2023conceptgraphsopenvocabulary3dscene41}} & LLava-7b-v0                                     & 13.3                        & 6.2           & 16.0                        & 7.2           \\
                               & LLava-7b-v1.5                                   & 16.9                        & 9.4          & 24.9                        & 15.4          \\
                               & LLava-7b-v1.6                                   & 18.4                        & 12.8          & 27.2                        & 17.7          \\ \hline
\multirow{3}{*}{BBQ~\cite{linok2024barequeriesopenvocabularyobject42}}           & LLava-7b-v0                                     & 27.2                        & 17.9          & 31.4                        & 26.3          \\
                               & LLava-7b-v1.5                                   & 31.7                        & 20.3          & 39.5                        & 30.6          \\
                               & LLava-7b-v1.6                                   & 34.2                        & 22.7          & 41.3                        & 33.5          \\ \hline
\multirow{3}{*}{Ours}          & LLava-7b-v0                                     &60.2                       & 45.4        & 50.6                      &42.1          \\
                               & LLava-7b-v1.5                                   & 60.9                      & 46.0         &51.1                      & 42.8         \\
                               & \textbf{LLava-7b-v1.6}                          & \textbf{61.1}               & \textbf{46.3} & \textbf{51.3}               & \textbf{43.5} \\ \hline
\end{tabular}
}
\label{llava}
\end{table}

\begin{figure}[!htp]
\centering
\includegraphics[width=1\linewidth]{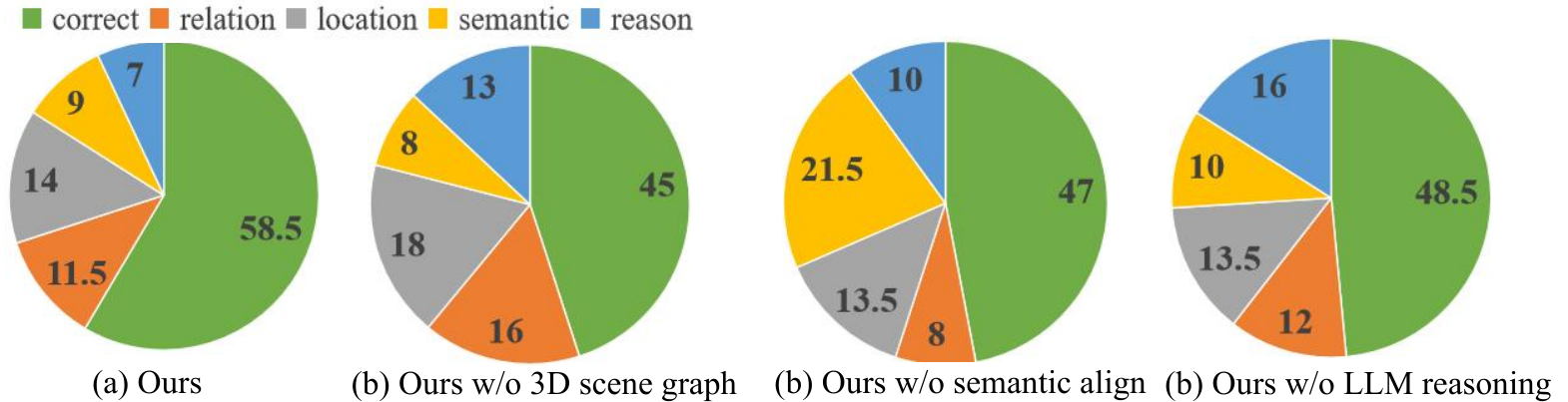}
\caption{Error analysis of FreeQ-Graph on  ScanRefer dataset.}
\label{error}
\end{figure}
\subsubsection{Semantic  Alignment}
%We also conduct ablation experiments on Semantic Point-node Alignment, 
As shown in rows 2 and 3 of Table ~\ref{ablation}, Aligning the semantic features of the graph nodes with the semantic consistent superpoint features significantly enhances the performance of two datasets. This highlights that the proposed module effectively aligns the consistent semantic label of the graph nodes, as shown in Fig.~\ref{graph}.
\subsubsection{LLM-based CoT-reasoning}
Rows 3-4 in Table~\ref{ablation} show that the LLM-based reasoning enhances the model's ability to infer complex semantics.
It indicates that integrating scene-level and object-level information fosters a more nuanced understanding of complex scenes. Furthermore, by decomposing the complex query into two stages, the model more effectively identifies candidate objects and their relationships, enabling deeper analysis to determine the final target.

\subsubsection{Reasoning algorithms}
We explored how reasoning algorithms affect 3D object grounding on the Sr3D~\cite{achlioptas2020referit3d} and Nr3D\cite{achlioptas2020referit3d}, evaluating with Acc@0.1 and Acc@0.25 metrics. As shown in Table~\ref{reasoning}, our model outperforms various SOTA reasoning methods. models~\cite{linok2024barequeriesopenvocabularyobject42,gu2023conceptgraphsopenvocabulary3dscene41}.  Moreover, our reasoning algorithm can seamlessly integrate with others, such as ConceptGraph~\cite{gu2023conceptgraphsopenvocabulary3dscene41}, significantly enhancing their ability to handle free-form complex semantic queries. It demonstrates the superiority of our LLM-based reasoning algorithm for free-form scene semantic queries.

\subsubsection{Computational costs}
As shown in Table~\ref{time}, our method achieves superior efficiency with significantly lower computational cost than other zero-shot, LLM-based approaches that require no pre-training or fine-tuning~\cite{gu2023conceptgraphsopenvocabulary3dscene41,linok2024barequeriesopenvocabularyobject42}. Unlike fully-supervised or fine-tuned LLM-based models that demand hours of training, our training-free framework highlights strong practical efficiency.
While we also use a GPT-based model for reasoning, like ConceptGraph and BBQ, our method delivers faster inference under the same settings. This is enabled by our semantically aligned 3D scene graph, which ensures accurate and efficient semantic representation and relation extraction. Furthermore, our CoT reasoning decomposes complex queries into manageable steps, improving reasoning speed. In contrast, ConceptGraph and BBQ rely solely on LLM outputs, often overlooking inconsistencies that lead to semantic misalignment and slower performance.

\subsubsection{Error Analysis} To assess reliance, we perform an error analysis on 200 randomly selected ScanRefer~\cite{chen2020scanrefer} samples (Fig.~\ref{error}), categorizing errors into 5 cases. Our scene graph enhances localization and relation detection, while semantic alignment reduces mislabeling errors. Our reasoning module effectively mitigates inference errors and keep stability.

%While both methods excel in descriptive queries, the CLIP-based approach falters on complex affordance and negation queries with ambiguous or semantically similar targets, whereas the LLM-based method uses advanced reasoning to resolve such issues effectively.
%While both methods perform comparably on descriptive queries, the CLIP-based approach struggles with more intricate affordance and negation queries, where targets lack direct semantic references or contain numerous semantically similar objects. In contrast, the LLM-based grounding method leverages advanced reasoning capabilities to more effectively discern targets with overlapping semantic features.

\subsubsection{Different LLaVA models.}
As shown in Table~\ref{llava}, ablation on different LLaVAs over Nr3D and Sr3D shows that advanced LLaVA improves grounding by reducing caption errors. Our model remains more stable than BBQ and ConceptGraph, indicating that our semantically consistent 3D scene graph and reasoning reduce reliance on specific  versions.

\section{Conclusion}
In this paper, we propose FreeQ-Graph, which enables free-form querying for 3D scene understanding. The main contribution lies in semantically consistent 3D scene graphs, which capture 3D semantic-aligned features.
Besides, we develop a LLM-based analysis and reasoning algorithm for free-form semantic querying. 
Experiments on 6 datasets show that our model excels in free-form queries and relational reasoning.
%Our model not only achieves superior performance in 3D open-vocabulary semantic segmentation and object grounding tasks, but it also accurately understands complex semantic queries.
Codes and datasets will be released.

\noindent \textbf{Limitations.} We consider potential limitations: 1) The use of multiple LLM and LVLM agents may increase inference time and costs. 2) Node captions are prone to errors due to the inherent limitations of current LVLMs. %We discuss the ablation study of different LLAVA models in Appendix E.

\tiny

%\begin{thebibliography}{1}
\bibliographystyle{IEEEtran}
\small
\bibliography{ref.bib}

\vfill

\end{document}